\documentclass[lettersize,journal]{IEEEtran}

\usepackage{amsmath,amsfonts,amssymb}
\usepackage{algorithm}
\usepackage{algorithmic}
\usepackage{array}
\usepackage{booktabs}
\usepackage[caption=false,font=footnotesize,labelfont=rm,textfont=rm]{subfig}
\usepackage{float}
\usepackage{textcomp}
\usepackage{stfloats}
\usepackage{capt-of}
\usepackage{url}
\usepackage{graphicx}
\usepackage{xcolor}
\usepackage{pifont}
\usepackage{enumitem}
\usepackage{cite}
\usepackage{soul}
\usepackage{multirow}
\usepackage{makecell}
\usepackage[nohyperlinks]{acronym}
\acrodef{capql}[CAPQL]{Concave Augmented Pareto Q-learning}
\acrodef{cot}[CoT]{Cost of Transport}
\acrodef{dof}[DoF]{Degrees of Freedom}
\acrodef{dr}[DR]{Domain Randomization}
\acrodef{hv}[HV]{Hypervolume}
\acrodef{mae}[MAE]{Mean Absolute Error}
\acrodef{sorl}[SORL]{Single-Objective Reinforcement Learning}
\acrodef{moppo}[MOPPO]{Multi-Objective Proximal Policy Optimization}
\acrodef{dpmorl}[DPMORL]{Distributional Pareto-Optimal Multi-Objective Reinforcement Learning}
\acrodef{crps}[CRPS]{Continuous Ranked Probability Score}
\acrodef{cvar}[CVaR]{Conditional Value-at-Risk}
\acrodef{gae}[GAE]{Generalized Advantage Estimation}
\acrodef{gpu}[GPU]{Graphics Processing Unit}
\acrodef{mdp}[MDP]{Markov Decision Process}
\acrodef{momdp}[MOMDP]{Multi-Objective Markov Decision Process}
\acrodef{morl}[MORL]{Multi-Objective Reinforcement Learning}
\acrodef{ood}[OOD]{Out-of-Distribution}
\acrodef{pcn}[PCN]{Preference-Conditioned Network}
\acrodef{pdmorl}[PD-MORL]{Preference-Driven Multi-Objective Reinforcement Learning}
\acrodef{pd}[PD]{Proportional-Derivative}
\acrodef{ppo}[PPO]{Proximal Policy Optimization}
\acrodef{promo}[PROMO]{\textbf{Pr}eference-C\textbf{o}nditioned \textbf{M}ulti-\textbf{O}bjective Reinforcement Learning for Quadrupedal Robots}
\acrodef{rl}[RL]{Reinforcement Learning}
\acrodef{rma}[RMA]{Rapid Motor Adaptation}
\acrodef{tar}[TAR]{Teacher-Aligned Representations}
\acrodef{td}[TD]{Temporal-Difference}
\acrodef{uvfa}[UVFA]{Universal Value Function Approximator}

\newif\ifreviewbuild
\reviewbuildtrue   

\makeatletter
\@ifundefined{acr@used}{\newif\ifacr@used}{}
\makeatother

\graphicspath{{figures/}{figures/sim/}{figures/hw/}{figures/train/}{figures/ablation/}}

\setlist{nosep,leftmargin=*}
\makeatletter
\def\section{\@startsection{section}{1}{\z@}{1.1ex plus 0.3ex minus 0.2ex}{0.7ex plus 0.2ex}{\normalfont\large\bfseries}}
\def\subsection{\@startsection{subsection}{2}{\z@}{0.9ex plus 0.25ex minus 0.15ex}{0.45ex plus 0.15ex}{\normalfont\normalsize\bfseries}}
\let\addcontentsline\@gobblethree
\makeatother

\newcounter{proposition}
\renewcommand{\theproposition}{\arabic{proposition}}
\newenvironment{proposition}[1][]{%
  \refstepcounter{proposition}%
  \par\addvspace{0.8\baselineskip}%
  \noindent\textbf{Proposition~\theproposition.}%
  \if\relax\detokenize{#1}\relax\else\ \textit{#1.}\fi\quad\ignorespaces%
}{%
  \par\addvspace{0.4\baselineskip}%
}

\begin{document}

\title{PROMO: Preference-conditioned Multi-Objective Reinforcement Learning for Quadrupedal Robots}

\author{Amr~Mousa\textsuperscript{1},
Rifny~Rachman\textsuperscript{1},
Neil~Karavis\textsuperscript{2},
Michele~Caprio\textsuperscript{1,3},
and~Richard~Allmendinger\textsuperscript{1}%
\thanks{\textsuperscript{1}The University of Manchester, United Kingdom.}%
\thanks{\textsuperscript{2}BAE Systems, United Kingdom.}%
\thanks{\textsuperscript{3}University of Warwick, United Kingdom.}%
\thanks{Project website, code, and videos: \protect\url{https://amrmousa.com/promo/}.%
\par\vspace{0.4\baselineskip}\leftskip=1em\noindent
\textit{This work has been submitted to the IEEE for possible publication. Copyright may be transferred without notice, after which this version may no longer be accessible.}}}


\IEEEaftertitletext{%
\begin{center}
\stepcounter{figure}
\includegraphics[width=\textwidth]{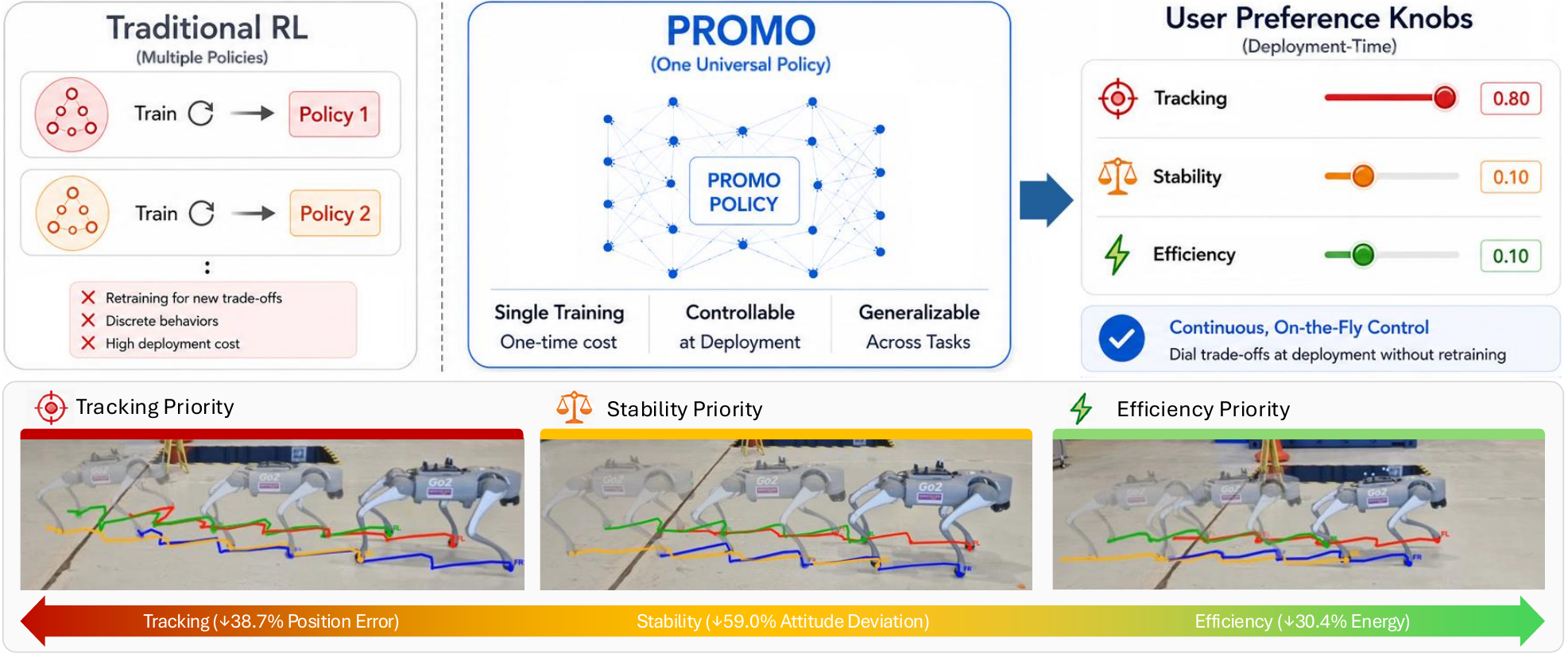}
\end{center}
\vspace{-0.15\baselineskip}
}

\maketitle

\begin{abstract}
Quadrupedal locomotion requires balancing conflicting objectives such as command tracking, stability, and energy efficiency, yet conventional reinforcement learning (RL) hard-codes these priorities into a fixed scalar reward at training time. 
We present \textbf{PROMO} (Preference-Conditioned Multi-Objective Reinforcement Learning), a semantic multi-objective framework that makes this trade-off an explicit runtime input to a single locomotion policy. PROMO conditions the policy on deployment-facing preferences while keeping embodiment-specific locomotion priors fixed, thereby separating operator intent from reward-shaping terms required for viable gait generation. 
Compared with fixed-objective controllers, multi-objective baselines, and independently trained specialists, PROMO achieves objective specialization and robustness from a single deployable policy. 
Across $100$ sampled preferences in simulation, $67$ behaviors are non-dominated under exact Pareto dominance, with a mean preference–objective correlation of $0.843$, demonstrating broad Pareto coverage and predictable preference response. 
The same policy transfers zero-shot to a Unitree Go2, where preference changes alone reduce specific energy by up to $30.4\%$, position error by $38.7\%$, and peak body-attitude deviation by $59.0\%$ relative to the balanced preference. 
These results establish preference-conditioned multi-objective RL as a practical runtime interface for adaptive legged locomotion, extending its role beyond offline Pareto-set construction.
Open-source code and videos are available at \url{https://amrmousa.com/promo/}.
\end{abstract}


\acresetall

\section{Introduction}

\IEEEPARstart{R}{obotic} locomotion via \ac{rl} has moved quadrupeds from controlled demonstrations toward agile and deployable systems~\cite{kumar2021rma,rudin2022learning,miki2022learning}. Most learned controllers, however, optimize a scalar reward that fixes the trade-off among command tracking, stability, energy efficiency, and locomotion regularization. This is restrictive in long-horizon autonomy, where terrain, payload, mission urgency, operator intent, or hardware state can change which objective should dominate.

After training, a robot can receive new motion commands, but the objective trade-off governing \emph{how} those commands are executed remains locked. Changing that trade-off usually requires a reward redesign, policy retraining, or controller switching. The deployment challenge is therefore to learn robust locomotion while making the underlying objective priorities selectable after training.

\ac{morl} provides a principled basis for adapting trade-offs by representing distinct objectives through vector-valued rewards and optimizing their utility under a preference~\cite{roijers2013survey,hayes2022practical,radulescu2020utility}. However, directly treating locomotion reward terms as independent objectives creates a fundamental mismatch with learned locomotion. Robotic rewards typically combine many shaping terms for posture, contact timing, smoothness, joint limits, and other requirements for viable motion~\cite{rudin2022learning,miki2022learning,mousa2025tar}. Making each term an objective yields a high-dimensional Pareto space that becomes harder to optimize as dimensionality grows~\cite{ALLMENDINGER2022105857}, while exposing low-level coefficients that do not express behavioral intent.

This objective-definition problem is central to deployable robotic \ac{morl}: the optimized dimensions need not coincide with the components used to shape a robust controller. We therefore separate a compact semantic objective vector—tracking, stability, and efficiency—from fixed locomotion priors. This keeps embodiment-specific requirements internal to the controller and makes the preference vector a meaningful runtime interface rather than a reformulation of the full reward function.

Behavior-conditioned locomotion addresses flexibility through a different interface. Methods such as Walk These Ways~\cite{margolis2023walk} expose gait descriptors that specify \emph{how} the robot should move through morphology, timing, or stance parameters. Preference-conditioned control instead specifies \emph{what} the robot should prioritize, leaving the physical gait realization to the policy. This distinction matters for deployment: semantic preferences should retain their meaning across commands, disturbances, and operating conditions rather than become handcrafted gait-tuning variables.

These observations motivate \ac{promo}, a preference-conditioned \ac{morl} framework that turns a fixed reward trade-off into a runtime \emph{semantic behavior interface}. A single policy is conditioned on preferences over three deployment-facing objectives—tracking, stability, and efficiency—while embodiment-specific locomotion priors remain fixed. 

The resulting preference vector is both a deployment-time control input and an interpretable representation of behavioral intent. Rather than manipulating latent gait codes or implementation-level reward weights, the operator specifies \emph{why} behavior should change—for example, by prioritizing stability over tracking—and the policy learns \emph{how} to realize that trade-off.

We evaluate whether this interface is controllable in simulation and remains preference-responsive after zero-shot transfer to hardware. In simulation, \ac{promo} remains competitive at the balanced operating point while providing tracking, stability, and efficiency specialization from a single policy, with lower failure rates than the evaluated preference-conditioned and specialist baselines. A dense 100-preference study shows that these anchor behaviors lie within a broader controllable interface. On a Unitree Go2, changing only the preference produces large, directionally consistent changes in tracking, stability, and energy efficiency across five trials per condition. To our knowledge, this is the first real demonstration of semantic preference-conditioned \ac{morl} on a robotic platform, moving \ac{morl} from offline Pareto optimization toward an operator-facing mechanism for adaptive robot behavior.

Our \textbf{core contributions} are:
\begin{enumerate}
    \item[\textbf{C1}] \textbf{A semantic locomotion interface.}
    We formulate runtime-adjustable quadruped control as preference-conditioned \ac{morl} over tracking, stability, and efficiency, making the objective trade-off an explicit deployment-time input that represents operator intent.

    \item[\textbf{C2}] \textbf{Objective/prior factorization for controllable locomotion.}
    We address the mismatch between vector-valued \ac{morl} objectives and dense robotic reward design by separating operator-facing semantic objectives from fixed embodiment-level locomotion priors. Ablations show that this factorization is critical for controllability and reliability: the fixed-prior formulation attains a $6\%$ fall rate under challenging evaluation conditions, whereas alternative designs increase it to as much as $21.8\%$ and degrade preference controllability.

    \item[\textbf{C3}] \textbf{Zero-shot semantic \ac{morl} on a real quadruped.}
    We provide, to our knowledge, the first real-robot demonstration of semantic preference-conditioned \ac{morl} for locomotion. A single policy transfers zero-shot to a Unitree Go2 and modulates tracking, stability, and efficiency through preference alone. Relative to a balanced configuration, preference modulation reduces specific energy by up to $30.4\%$, position error by up to $38.7\%$, and peak body-attitude deviation by up to $59.0\%$.

    \item[\textbf{C4}] \textbf{Open-source implementation and complete experimental suite.}
    We release the implementation of \ac{promo} together with all evaluated baselines, experimental tasks, evaluation protocols, and ablation studies, enabling reproduction of the complete experimental pipeline and facilitating future benchmarking and extensions.

\end{enumerate}

The remainder of this paper is organized as follows. Section~\ref{sec:related_work} reviews related work. Section~\ref{sec:promo} formulates \ac{promo} and describes its training and deployment. Sections~\ref{sec:experimental_evaluation} and~\ref{sec:real_robot} evaluate the framework in simulation and on the Unitree Go2 robot, respectively. Section~\ref{sec:discussion} discusses the implications and limitations, and Section~\ref{sec:conclusion} concludes the paper.

\section{Related Work}
\label{sec:related_work}

This section positions \ac{promo} relative to learned quadruped locomotion, reward design and runtime behavioral interfaces, preference-conditioned \ac{morl}, and the use of Pareto coverage for behavioral control.

\subsection{Learned Quadruped Locomotion}

\ac{rl}-based locomotion has made quadruped control increasingly robust to sim-to-real transfer, partial observability, and environmental uncertainty. Actuator modeling~\cite{hwangbo2019learning}, domain randomization~\cite{tan2018sim2real}, privileged adaptation~\cite{kumar2021rma}, massively parallel training~\cite{rudin2022learning}, perception-aware control~\cite{miki2022learning,nahrendra2023dreamwaq}, implicit adaptation~\cite{long2024him,long2025hinf}, and teacher-aligned representations~\cite{mousa2025tar} have enabled reliable deployment under challenging conditions. Most controllers nevertheless remain \ac{sorl} policies: commands and observations may change at runtime, but the reward-level priority among tracking, stability, efficiency, and regularization is fixed.

This raises a question: \emph{which parts of the locomotion objective should remain fixed, and which should be exposed for runtime adaptation?}

\subsection{Reward Design and Runtime Behavioral Interfaces}

Modern locomotion rewards mix task objectives with posture regulation, contact shaping, actuator smoothness, joint-limit penalties, and other embodiment-specific terms~\cite{rudin2022learning,miki2022learning,mousa2025tar}. These dense components suppress different failure modes and collectively shape viable motion, but they do not all represent objectives over which an operator should express a preference. Treating contact regularity, posture, or action smoothness as independent objectives would expose reward-engineering decisions as deployment controls and enlarge the preference space whenever another shaping term is introduced. Reward shaping~\cite{ng1999shaping}, constrained \ac{rl}~\cite{achiam2017cpo}, constrained \ac{morl}~\cite{gu2025safeBalanced}, and inverse reward design~\cite{hadfieldmenell2017ird} similarly distinguish task objectives from auxiliary signals or constraints. \ac{promo} operationalizes this distinction through semantic objective/prior factorization: task-level trade-offs define the preference vector, whereas embodiment-level requirements remain fixed locomotion priors.

Runtime flexibility can also be achieved through behavior conditioning~\cite{margolis2023walk, wu2023multipleGaits, lee2026gaitParameterized}. These methods show that one policy can span diverse behaviors through operator-controlled gait variables, but the interface primarily specifies \emph{how} the robot should move through morphology or timing. Semantic preference conditioning instead specifies \emph{what} the robot should prioritize.

This shifts the problem from commanding behaviors to trading off objectives: \emph{how can a single policy realize a continuum of semantic preferences?}

\subsection{MORL Preference Conditioning}

\ac{morl} formalizes control with multiple conflicting objectives through vector-valued returns~\cite{roijers2013survey}, scalarization and utility theory~\cite{radulescu2020utility}, and Pareto optimality~\cite{hayes2022practical}. The algorithmic literature includes Pareto-front approximation~\cite{xu2020pgmorl}, decomposition~\cite{basaklar2023pdmorl}, constrained optimization~\cite{lu2023capql}, diversity-driven coverage~\cite{ambadkar2026d3po}, and preference-conditioned policy learning~\cite{liu2025cmorl,liu2025pslmorl}. Standardized frameworks such as MO-Gymnasium and MORL-Baselines support systematic comparison~\cite{felten2023morlToolkit,felten2024morld}.

Preferences may be introduced a priori, a posteriori, or interactively, and expressed through weights, reference points, or solution comparisons~\cite{miettinen1999nonlinear}. A fixed a priori weight vector yields one scalarized solution; \ac{promo} instead learns a weight-conditioned approximation of the supported Pareto front, enabling a posteriori behavior selection at deployment.

Preference-conditioned single-policy methods are attractive for deployment because one policy can represent a continuum of trade-offs. They are based on universal value function approximation~\cite{schaul2015uvfa} and reward-conditioned \ac{rl}~\cite{kumar2019rewardconditioned,reymond2022pcn}. In our evaluation, \ac{moppo} provides a single-policy preference-conditioned \ac{ppo} baseline~\cite{terekhov2024moppo}, whereas \ac{dpmorl} represents a multi-policy route to Pareto-front approximation~\cite{cai2023dpmorl}.

Robotics applications remain comparatively limited, including prediction-guided multi-objective continuous control~\cite{xu2020pgmorl}, constrained robot control~\cite{huang2021lp3}, and gradient-conflict resolution~\cite{munn2025gcrppo}. These works establish the relevance of \ac{morl} to robotics, but do not address how preference-conditioned optimization should be formulated and deployed for contact-rich legged robots.

This leaves an interface-level question: \emph{does learning a broad Pareto set make preferences useful as deployment-time controls?}

\subsection{From Pareto Coverage to Behavioral Control}

Standard \ac{morl} benchmarks ignore failures, dense reward structure, embodiment-specific priors, sim-to-real transfer, and hardware deployment. Strong Pareto performance therefore does not necessarily imply that preferences provide a reliable control interface on a physical robot.

Hypervolume is the principal Pareto-front approximation metric used in our study. It is Pareto compliant and measures the objective-space volume dominated by an approximation set and bounded by a reference point, summarizing convergence and diversity~\cite{knowles2006tutorial}. However, it does not establish whether changing a preference predictably changes behavior. Recent work therefore considers preference sensitivity, monotonicity, objective alignment, and behavioral consistency~\cite{yang2025pcrl,delasheras2026controllability,jiang2026graphallocbench}. A policy may approximate a broad Pareto front while still providing an unreliable interface if preferences map inconsistently to realized behaviors.

For physical deployment, Pareto quality must be evaluated together with preference controllability, robustness, and semantic interpretability. \ac{promo} combines semantic objective specification, preference-conditioned control, locomotion-specific priors, interface-oriented evaluation, and physical deployment in a single framework.

\section{Methodology: Semantic Preference-Conditioned Locomotion}
\label{sec:promo}
\label{sec:problem_formulation}

This section formulates \ac{promo} and describes its architecture, semantic objective factorization, preference-conditioned optimization, and mechanisms for responsive deployment.

\subsection{Problem Formulation and Semantic Interface}

We model locomotion as a partially observed multi-objective decision process. Let $o_t$ denote the current deployable observation, including proprioception and the velocity command, let $o_{t-H:t}$ denote its history, and let $s_t^{\mathrm{priv}}$ collect the corresponding simulator-only state used during training. The preference is $w_t\in\Delta^{K-1}$. The history encoder $E_H$ and velocity estimator produce
\begin{equation}
z_t^H=E_H(o_{t-H:t},w_t),
\qquad
\hat v_t^b=E_V(z_t^H,o_{t-4:t}),
\label{eq:history_velocity_def}
\end{equation}
where $\hat v_t^b$ estimates the three-dimensional base linear velocity. The deployed policy outputs $a_t$ according to $\pi_\theta(a_t\mid o_t,z_t^H,\hat v_t^b,w_t)$. The stop-gradient operations used during optimization are stated in Section~\ref{sec:architecture_overview}. For quadrupedal locomotion, the semantic reward vector is
\begin{equation}
\mathbf{r}_t^{\mathrm{sem}} = \bigl[
r_t^{\mathrm{track}},
r_t^{\mathrm{stab}},
r_t^{\mathrm{eff}}
\bigr]^\top,
\end{equation}
where the objectives correspond to command tracking, body stability, and energetic efficiency. Let $\mathcal{I}=\{\mathrm{track},\mathrm{stab},\mathrm{eff}\}$ denote the semantic objective index set, with $K=|\mathcal{I}|=3$, and let $r_{i,t}$ denote component $i\in\mathcal{I}$ of $\mathbf{r}_t^{\mathrm{sem}}$. Semantic utility is defined as
\begin{equation}
U(\mathbf{r}_t^{\mathrm{sem}},w_t) = w_t^\top \mathbf{r}_t^{\mathrm{sem}}.
\end{equation}
During training, semantic utility is augmented by a fixed locomotion prior,
\begin{equation}
u_t(w_t) = w_t^\top \mathbf{r}_t^{\mathrm{sem}} + \lambda_{\mathrm{prior}} r_t^{\mathrm{prior}},
\end{equation}
where $r_t^{\mathrm{prior}}$ contains embodiment-level regularization terms shared across all preferences and $\lambda_{\mathrm{prior}}$ is its fixed coefficient. For the discount factor $\gamma$, the semantic and prior returns are
\begin{equation}
J_{\mathrm{sem}}(\pi_\theta,w)
=
\mathbb{E}_{\pi_\theta,w}
\left[
\sum_{t\ge0}\gamma^t\mathbf{r}_t^{\mathrm{sem}}
\right],
\label{eq:semantic_return_def}
\end{equation}
\begin{equation}
J_{\mathrm{prior}}(\pi_\theta,w)
=
\mathbb{E}_{\pi_\theta,w}
\left[
\sum_{t\ge0}\gamma^t r_t^{\mathrm{prior}}
\right].
\label{eq:prior_return_def}
\end{equation}

Because all semantic objectives are maximized, we use the following Pareto terminology~\cite{ALLMENDINGER2022105857,hayes2022practical}. Let $\Pi$ be the feasible policy class and let $\mathbf{J}(\pi)\in\mathbb{R}^{K}$ denote the semantic return vector of $\pi\in\Pi$ as defined in Eq.~\eqref{eq:semantic_return_def}. A policy $\pi^a$ Pareto dominates $\pi^b$, written $\pi^a\succ_P\pi^b$, if
\begin{equation}
\begin{aligned}
J_i(\pi^a)&\geq J_i(\pi^b) && \forall i\in\mathcal I,\\
J_j(\pi^a)&>J_j(\pi^b) && \text{for some }j\in\mathcal I.
\end{aligned}
\label{eq:pareto_dominance}
\end{equation}
A policy is \emph{Pareto optimal}  if no feasible policy dominates it. The \emph{Pareto set} $\mathcal P\subseteq\Pi$ contains these policies in the policy space, whereas the \emph{Pareto front} $\mathcal F=\{\mathbf J(\pi):\pi\in\mathcal P\}$ contains their return vectors in the semantic objective space. Within a finite evaluated set, a policy is \emph{non-dominated} if no other evaluated policy dominates it; this finite set is therefore a sampled approximation to, rather than proof of, the unknown true Pareto front. For the learned conditional actor, each preference induces $\pi_w(a\mid o,z^H,\hat v^b)=\pi_\theta(a\mid o,z^H,\hat v^b,w)$ and the corresponding outcome $\mathbf J(\pi_w)=J_{\mathrm{sem}}(\pi_\theta,w)$.

A single preference is sampled per environment from the truncated-simplex distribution and kept fixed within each rollout. Preference changes occur only during deployment or evaluation. Implementation details are provided in Section~\ref{sec:platform_training}.

The semantic objective space contains only deployment-facing dimensions whose relative importance is specified by the user. Embodiment-level regularizers, including contact timing, contact balance, and posture constraints, are represented by the fixed prior $r_t^{\mathrm{prior}}$. This factorization prevents dense reward components from becoming artificial objective dimensions. Thus, $w$ represents behavioral intent rather than implementation-level reward weights, and Pareto and preference-controllability analyses are defined exclusively over $\mathbf{r}_t^{\mathrm{sem}}$.

At deployment, a human operator, planner, or supervisory controller supplies $w_t\in\Delta^{K-1}$ to the fixed policy $\pi_\theta(a_t\mid o_t,z_t^H,\hat v_t^b,w_t)$. In all reported evaluations, each preference is kept fixed throughout a rollout, directly associating the realized behavior with the requested semantic trade-off. Appendix~\ref{app:rewards_preferences} gives the full reward decomposition.

\subsection{Architecture Overview}
\label{sec:architecture_overview}

\begin{figure*}[t]
\centering
\includegraphics[width=1.0\linewidth]{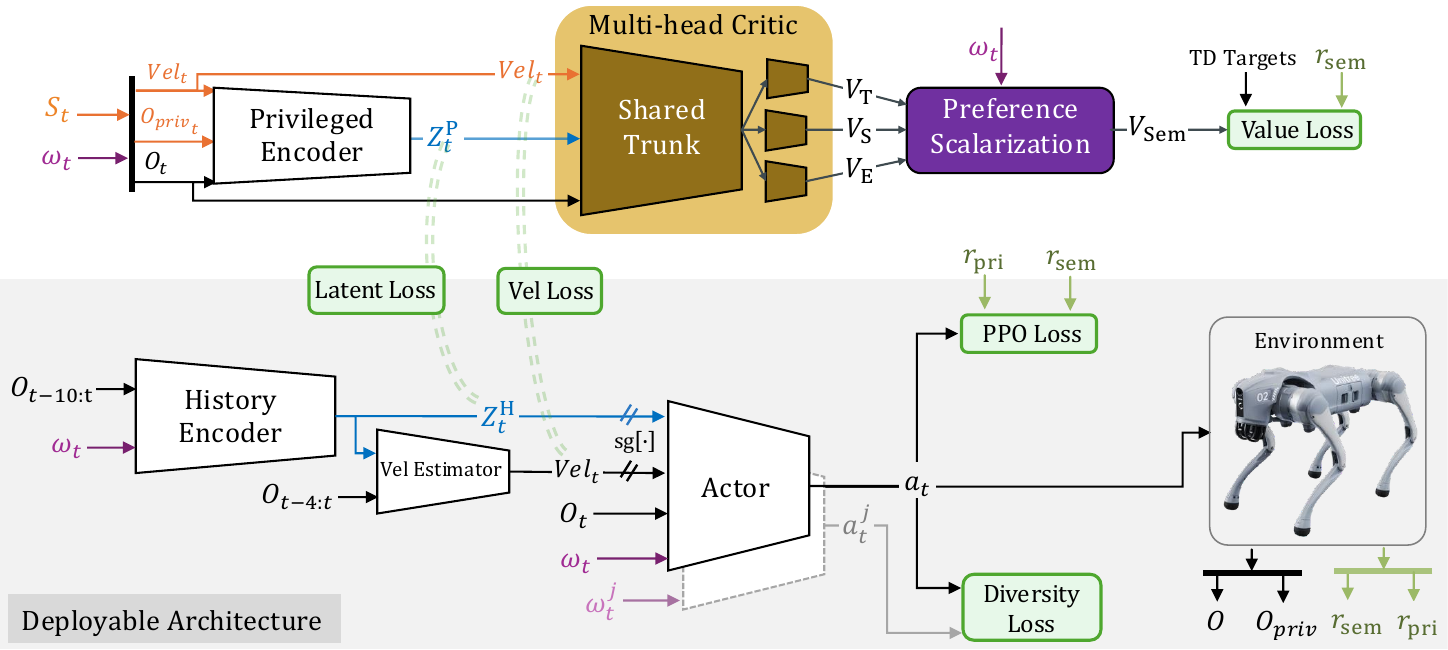}
\caption{\acs{promo} architecture and training loop. The privileged latent $z_t^P$ enters the critic, whereas the history latent $z_t^H$ and estimated velocity $\hat v_t^b$ enter the actor through stop-gradient operations. Both encoders receive $w_t$. The semantic critic values are scalarized by $w_t$ before the clipped \acs{ppo} update. The shaded lower path contains the modules retained at deployment: the history encoder, velocity estimator, and actor; the privileged encoder and critic are discarded.}
\label{fig:architecture}
\end{figure*}

\ac{promo} extends the locomotion architecture of \ac{tar}~\cite{mousa2025tar} to preference-conditioned \ac{morl} with semantic objectives. Its privileged encoder $E_P$ and history encoder $E_H$ produce distinct latents:
\begin{equation}
z_t^P=E_P(s_t^{\mathrm{priv}},w_t),
\qquad
z_t^H=E_H(o_{t-H:t},w_t).
\label{eq:encoder_latents}
\end{equation}
Here, $s_t^{\mathrm{priv}}$ comprises the current deployable observation, ground-truth base velocity, and the additional privileged simulator signals. Both encoders are preference-conditioned, as shown in Fig.~\ref{fig:architecture}. During training, the critic receives $(o_t,v_t^b,z_t^P)$ and predicts one value for each semantic objective. It does not receive $z_t^H$; the preference enters its representation through $z_t^P$ and is applied explicitly when the objective-wise values are scalarized. In contrast, the actor receives $(o_t,z_t^H,\hat v_t^b,w_t)$ and outputs $12$ joint-position targets.

The history representation is aligned with a detached privileged target using
\begin{equation}
\mathcal L_{\mathrm{align}}
=
\mathbb E_t\!\left[
\left\lVert z_t^H-\operatorname{sg}[z_t^P]\right\rVert_2^2
\right],
\label{eq:latent_alignment_loss}
\end{equation}
where $\operatorname{sg}[\cdot]$ denotes stop-gradient. The critic loss updates the privileged encoder, whereas Eq.~\eqref{eq:latent_alignment_loss} does not propagate into it. The velocity estimator uses the history latent and the latest five observations, as defined in Eq.~\eqref{eq:history_velocity_def}, and minimizes
\begin{equation}
\mathcal L_{\mathrm{vel}}
=
\mathbb E_t\!\left[
\left\lVert \hat v_t^b-v_t^b\right\rVert_2^2
\right].
\label{eq:velocity_estimator_loss}
\end{equation}
The history encoder is updated by $\lambda_{\mathrm{align}}\mathcal L_{\mathrm{align}}+\mathcal L_{\mathrm{vel}}$ with $\lambda_{\mathrm{align}}=1.0$, and the velocity estimator is updated by $\mathcal L_{\mathrm{vel}}$. Actor-side inputs use $\operatorname{sg}[z_t^H]$ and $\operatorname{sg}[\hat v_t^b]$, preventing policy gradients from entering either auxiliary module. The privileged encoder nevertheless receives gradients from the semantic critic loss through $z_t^P$.

Compared with \ac{tar}, \ac{promo} replaces the recurrent history encoder with a multilayer-perceptron encoder, removes the forward-dynamics auxiliary model, and introduces preference conditioning throughout the actor--critic architecture. The result is a lightweight architecture for preference-conditioned flat-terrain locomotion that retains privileged representation alignment. Figure~\ref{fig:architecture} summarizes training and deployment; network, simulator, and optimization details are provided in Appendix~\ref{app:training_sim_config}.

Algorithm~\ref{alg:promo_training} summarizes the resulting update paths. It separates rollout inference from the privileged quantities and auxiliary targets used only during optimization.
\begin{algorithm}[t]
\caption{\acs{promo} training and deployment}
\label{alg:promo_training}
\begin{algorithmic}[1]
\REQUIRE Actor $\pi_\theta$, critic $V_\psi$, encoders $E_P,E_H$, velocity estimator $E_V$
\FOR{each training rollout}
    \STATE Sample $w$ and hold it fixed for the rollout
    \FOR{each step $t$}
        \STATE $z_t^H\leftarrow E_H(o_{t-H:t},w)$; $\hat v_t^b\leftarrow E_V(z_t^H,o_{t-4:t})$
        \STATE $a_t\sim\pi_\theta(\cdot\mid o_t,\operatorname{sg}[z_t^H],\operatorname{sg}[\hat v_t^b],w)$
        \STATE Step the environment and store $o_t,s_t^{\mathrm{priv}},v_t^b,\mathbf r_t^{\mathrm{sem}},r_t^{\mathrm{prior}}$
    \ENDFOR
    \STATE $z_t^P\leftarrow E_P(s_t^{\mathrm{priv}},w)$ and evaluate $V_i^\psi(o_t,v_t^b,z_t^P)$ for all $i\in\mathcal I$
    \STATE Form semantic values, GAE advantages, $\mathcal L_{\mathrm{sem}}$, $\mathcal L_{\mathrm{PPO}}$, and $\mathcal L_{\mathrm{div}}^{\mathrm{act}}$
    \STATE Form $\mathcal L_{\mathrm{align}}=\lVert z_t^H-\operatorname{sg}[z_t^P]\rVert_2^2$ and $\mathcal L_{\mathrm{vel}}=\lVert\hat v_t^b-v_t^b\rVert_2^2$
    \STATE Update $(\psi,E_P)$ with $\mathcal L_{\mathrm{sem}}$
    \STATE Update $E_H$ with $\lambda_{\mathrm{align}}\mathcal L_{\mathrm{align}}+\mathcal L_{\mathrm{vel}}$
    \STATE Update $E_V$ with $\mathcal L_{\mathrm{vel}}$
    \STATE Update $\theta$ with $\mathcal L_{\mathrm{PPO}}+\lambda_{\mathrm{div}}\mathcal L_{\mathrm{div}}^{\mathrm{act}}$
\ENDFOR
\STATE \textbf{Deploy:} retain $(E_H,E_V,\pi_\theta)$; discard $(E_P,V_\psi)$
\end{algorithmic}
\end{algorithm}

\subsection{Semantic Objective Factorization and Value Learning}

Using the returns in Eqs.~\eqref{eq:semantic_return_def}--\eqref{eq:prior_return_def}, the training objective for preference $w_t$ is
\begin{equation}
J_{\mathrm{train}}(\theta,w_t)
=
w_t^\top J_{\mathrm{sem}}(\pi_\theta,w_t)
+
\lambda_{\mathrm{prior}}
J_{\mathrm{prior}}(\pi_\theta,w_t),
\label{eq:training_utility_main}
\end{equation}
where $\lambda_{\mathrm{prior}}=1$ in all experiments and is independent of $w_t$. The individual prior coefficients are listed in Appendix~\ref{app:rewards_preferences}. Pareto, controllability, and preference-response evaluations are computed only from $J_{\mathrm{sem}}$; the prior regularizes locomotion but is not exposed as a preference axis. Appendix~\ref{app:analysis} states the corresponding design property formally.

The semantic critic has one learned value head per semantic objective,
\begin{equation}
V^\psi_i(o_t,v_t^b,z_t^P),
\qquad i\in\mathcal{I},
\label{eq:scalar_critic_main}
\end{equation}
where $\psi$ denotes the critic parameters and $V^\psi_i$ approximates the expected discounted return for semantic objective $i$. The heads depend on the preference through the privileged latent $z_t^P=E_P(s_t^{\mathrm{priv}},w_t)$, but estimate only semantic returns. For the terminal indicator $d_t$, the one-step \ac{ppo} critic target is
\begin{equation}
y_{i,t}
=
r_{i,t}
+
\gamma(1-d_t)
V^{\psi}_{i}(o_{t+1},v_{t+1}^b,z_{t+1}^P),
\label{eq:critic_target_main}
\end{equation}
and the semantic critic minimizes the objective-wise regression loss
\begin{equation}
\mathcal L_{\mathrm{sem}}(\psi)
=
\frac{1}{K}
\sum_{i\in\mathcal{I}}
\left(
y_{i,t}
-
V^\psi_i(o_t,v_t^b,z_t^P)
\right)^2.
\label{eq:semantic_critic_loss_main}
\end{equation}
The fixed prior has no value head; see the objective/prior ablation in Appendix~\ref{app:analysis} and Table~\ref{tab:objprior_ablation}.

\subsection{Preference-Conditioned Policy Optimization}

For preference $w_t$, \ac{promo} scalarizes the semantic value heads as
\begin{equation}
V_{w_t}^\psi(o_t,v_t^b,z_t^P)
=
\sum_{i\in\mathcal{I}}
w_{t,i} V^\psi_i(o_t,v_t^b,z_t^P).
\label{eq:scalarized_value_main}
\end{equation}
For actor optimization, the semantic reward is scalarized as $r^{\mathrm{sem}}_t(w_t)=w_t^\top\mathbf{r}_t^{\mathrm{sem}}$. \ac{promo} then combines this reward with the fixed prior and the preference-scalarized semantic critic to estimate the actor advantage with \ac{gae} parameter $\lambda_{\mathrm{GAE}}$, where $\ell$ indexes future rollout steps:
\begin{align}
A_t
&=
\sum_{\ell\ge0}(\gamma\lambda_{\mathrm{GAE}})^\ell
\!\left(
\delta_{t+\ell}
\right),
\label{eq:promo_advantage_main}\\[5pt]
\delta_t
&=
r^{\mathrm{sem}}_t(w_t) 
+
\lambda_{\mathrm{prior}}r_t^{\mathrm{prior}}
\notag\\
&\quad+
\gamma(1-d_t)V_{w_{t+1}}^\psi
(o_{t+1},v_{t+1}^b,z_{t+1}^P)
\notag\\
&\quad-
V_{w_t}^\psi(o_t,v_t^b,z_t^P).
\notag
\end{align}
Because no prior value function is learned, the prior term enters this residual with a zero baseline; the only baseline subtraction is the scalarized semantic critic. Its contribution is therefore estimated through the $\lambda_{\mathrm{GAE}}$-discounted \ac{gae} trace. For $\lambda_{\mathrm{GAE}}=1$, this reduces to the Monte-Carlo prior return; for the setting used here ($0.95$), it is the standard biased, lower-variance \ac{gae} estimator rather than an unbiased estimate of the full prior return.

The actor update uses the advantage in Eq.~\eqref{eq:promo_advantage_main}. Define the complete actor input as $x_t^A=(o_t,\operatorname{sg}[z_t^H],\operatorname{sg}[\hat v_t^b],w_t)$. Then
\[
\varrho_t(\theta)
=
\frac{\pi_\theta(a_t\mid x_t^A)}
{\pi_{\theta_{\mathrm{old}}}(a_t\mid x_t^A)},
\qquad
\bar\varrho_t=\operatorname{clip}(\varrho_t,1-\epsilon,1+\epsilon),
\]
where $\theta_{\mathrm{old}}$ denotes the behavior-policy parameters used to collect the rollout and $\epsilon$ is the \ac{ppo} clipping width. The actor minimizes the negative clipped \ac{ppo} surrogate
\begin{equation}
\mathcal{L}_{\mathrm{PPO}}
=
-
\mathbb{E}_t
\left[
\min
\left(
\varrho_t A_t,
\bar\varrho_t A_t
\right)
\right].
\label{eq:ppo_objective_main}
\end{equation}
The optimizer constants are listed in Appendix~\ref{app:objective_algorithm}.

\subsection{Preference Responsiveness and Deployment}

Conditioning the actor on $w_t$ enables preference-dependent behavior, but does not guarantee visible use of the preference input. \ac{promo} therefore adds a small action-diversity regularizer. For two preferences $w_t$ and $w'_t$ evaluated at the same observation, detached history latent, and detached velocity estimate, let $\tilde x_t^A=(o_t,\operatorname{sg}[z_t^H],\operatorname{sg}[\hat v_t^b])$ and define
\[
\pi_t=\pi_\theta(\cdot\mid \tilde x_t^A,w_t),
\qquad
\pi'_t=\pi_\theta(\cdot\mid \tilde x_t^A,w'_t).
\]
The regularizer is
\begin{equation}
\mathcal{L}_{\mathrm{div}}^{\mathrm{act}}
=
-\mathbb{E}
\left[
\lVert w_t-w'_t\rVert
D_{\mathrm{sym}}(\pi_t,\pi'_t)
\right],
\label{eq:action_diversity_main}
\end{equation}
where $D_{\mathrm{sym}}$ is the symmetric KL divergence between the two diagonal-Gaussian action distributions,
\begin{equation}
D_{\mathrm{sym}}(\pi_t,\pi'_t)
=
\frac{1}{2}
\left[
D_{\mathrm{KL}}(\pi_t\Vert\pi'_t)
+
D_{\mathrm{KL}}(\pi'_t\Vert\pi_t)
\right],
\end{equation}
summed across action dimensions. The resulting actor objective is
\begin{equation}
\mathcal{L}_{\mathrm{actor}}
=
\mathcal{L}_{\mathrm{PPO}}
+
\lambda_{\mathrm{div}}
\mathcal{L}_{\mathrm{div}}^{\mathrm{act}},
\label{eq:actor_total_loss}
\end{equation}
with $\lambda_{\mathrm{div}}=10^{-5}$. Appendix~\ref{app:analysis} evaluates the coefficient sensitivity and the latent-diversity alternative.

At deployment, the history encoder $E_H$, velocity estimator $E_V$, and actor $\pi_\theta$ are retained. The privileged encoder, privileged state, ground-truth base velocity, and critic are discarded. The retained modules require only $(o_{t-H:t},w_t)$, and $w_t$ is the sole runtime mechanism for behavior modulation.

\section{Experimental Evaluation}
\label{sec:experimental_evaluation}

This section evaluates whether one preference-conditioned policy can provide a controllable semantic locomotion interface while retaining robust quadrupedal behavior. We organize the experiments around four questions:

\textbf{Q1: Training comparison.}
How does \ac{promo} optimize locomotion relative to single-objective controllers, existing \ac{morl} methods, and independently trained specialists under training metrics available to all methods? (Section~\ref{sec:baselines})

\textbf{Q2: Common semantic evaluation.}
Under common semantic metrics, does \ac{promo} preserve balanced locomotion quality and provide a useful multi-objective interface across requested behaviors? (Section~\ref{sec:baseline_semantic_eval})

\textbf{Q3: Preference-interface structure.}
For the deployed \ac{promo} policy, does varying the preference produce predictable semantic responses, and is Pareto coverage sufficient to characterize the interface? (Section~\ref{sec:promo_standalone_eval} and Appendix~\ref{app:analysis})

\textbf{Q4: Hardware deployment.}
Can the same preference-conditioned policy modulate physical locomotion through changes in priorities over the semantic objectives alone, and which failure modes appear on hardware? (Section~\ref{sec:real_robot})

\subsection{Setup and Baselines}
\label{sec:platform_training}

All policies are trained in Isaac Sim/IsaacLab with a $12$-\ac{dof} Unitree Go2 quadruped and $4096$ parallel flat-terrain environments.
Domain randomization covers friction, base mass and inertia, center-of-mass offsets, external perturbations, and joint-state initialization to support robustness and zero-shot hardware transfer.
The complete network, simulation, control, and domain-randomization configuration is provided in Appendix~\ref{app:training_sim_config}.

We compare \ac{promo} against three baseline groups:

\begin{itemize}
    \item \textbf{Single-objective locomotion controllers:}
    \ac{rma}~\cite{kumar2021rma} and \ac{tar}~\cite{mousa2025tar}, trained using the balanced semantic objective configuration. This group represents the default locomotion setting in which a single reward defines a fixed trade-off among conflicting objectives.

    \item \textbf{\ac{morl} methods:}
    \ac{moppo}~\cite{terekhov2024moppo}, a single-policy preference-conditioned baseline, and \ac{dpmorl}~\cite{cai2023dpmorl}, a multi-policy baseline that approximates the Pareto front using a set of specialized policies.

    \item \textbf{\ac{promo}-\ac{sorl}:}
    Three per-objective scalar specialists derived from the \ac{promo} architecture. They serve as a reference for the cost of amortizing multiple behaviors into a single preference-conditioned policy.
\end{itemize}

These comparisons test whether one \ac{promo} policy can provide controllable multi-objective locomotion while remaining competitive with established locomotion controllers, existing \ac{morl} approaches, and specialized scalar policies.
Appendix~\ref{app:baseline_implementations} summarizes the baseline implementations and shared controls used to keep the comparison methodologically consistent.

\subsection{Evaluation Metrics}

We evaluate training performance, semantic multi-objective performance, and preference controllability. Semantic evaluations use shared scripted commands and disturbances so preference-dependent differences are compared under matched rollout conditions; Fig.~\ref{fig:eval_protocol} illustrates the protocol, with full details in Appendix~\ref{app:evaluation_protocols}.

\textit{Training performance.}
We report the mean training reward and episode length over three independent seeds. These metrics characterize optimization under each method's native training objective and are therefore reported separately from the common semantic evaluation.
\textit{Semantic performance and robustness.}
All methods are evaluated on the same three semantic objectives: tracking, stability, and efficiency. The balanced evaluation reports semantic rewards, fall rate, and lower-tail body-attitude risk; the four-anchor comparison reports requested-objective scores, aggregate fall rate, and scalarized return. In Appendix~\ref{app:evaluation_protocols}, Table~\ref{tab:controllability}, lower-tail robustness is measured by \ac{cvar}$_{10}$ and Pareto quality by \ac{hv} after objective-range normalization.

For method/run $m$, let $\mathcal Y_m\subset\mathbb R^3$ denote its semantic outcomes. We define common empirical bounds from the union of run-level non-dominated sets,
\begin{equation}
\mathcal R=\bigcup_m\operatorname{ND}(\mathcal Y_m),\qquad
J_i^{\min/\max}=\min/\max_{\mathbf J\in\mathcal R}J_i,
\label{eq:hv_common_pool}
\end{equation}
and normalize all methods using
\begin{equation}
\hat J_i=\frac{J_i-J_i^{\min}}{J_i^{\max}-J_i^{\min}},\qquad
r_i=J_i^{\min}-0.05(J_i^{\max}-J_i^{\min}).
\label{eq:hv_objective_normalization}
\end{equation}
Since all objectives are maximized, the common reference maps to $\hat{\mathbf r}=(-0.05,-0.05,-0.05)$. We compute \ac{hv} from each normalized run-level non-dominated set in this common 3-D space. Only the objective ranges are normalized; \ac{hv} itself is not, giving a bounding-box upper bound of $1.05^3=1.157625$. Larger \ac{hv} indicates greater Pareto coverage.

\textit{Preference controllability.}
Beyond Pareto quality, controllability measures whether changing a preference produces the corresponding semantic response. For objective $i$, we compute $\rho_i=\operatorname{Spearman}(w_i,J_i^{\mathrm{sem}})$~\cite{spearman1904proof} and average across objectives. We distinguish four-anchor controllability for baseline comparison from dense controllability for the 100-preference \ac{promo} evaluation.

Finally, \ac{promo} is evaluated over $100$ preferences to characterize response trends, Pareto structure, and locomotion failures. Higher values are better for reward, utility, controllability, and range-normalized-objective \ac{hv}; lower values are better for fall rate and body-attitude tail risk.

\subsection{Training and Baseline Comparison}
\label{sec:sim_results}
The simulation evaluation establishes the interface properties needed before hardware deployment, rather than treating high training reward as sufficient evidence. The results combine optimization behavior, common semantic metrics, and a standalone analysis of the policy's controllability and Pareto coverage.

\begin{figure*}[t]
\centering
\subfloat[Locomotion baselines: reward]{\includegraphics[width=0.32\linewidth]{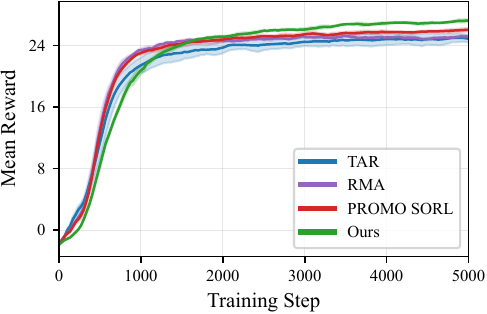}}\hfill
\subfloat[\acs{morl} baselines: reward]{\includegraphics[width=0.32\linewidth]{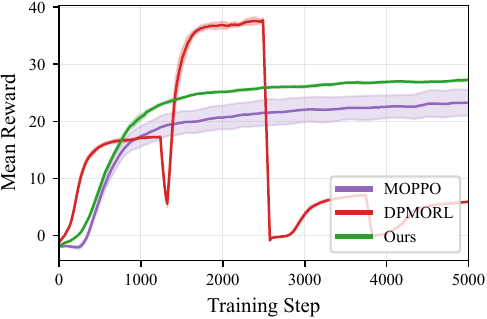}}\hfill
\subfloat[Specialists: reward]{\includegraphics[width=0.32\linewidth]{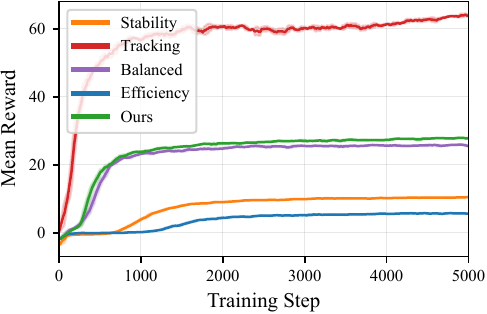}}\\[2pt]
\subfloat[Locomotion baselines: episode length]{\includegraphics[width=0.32\linewidth]{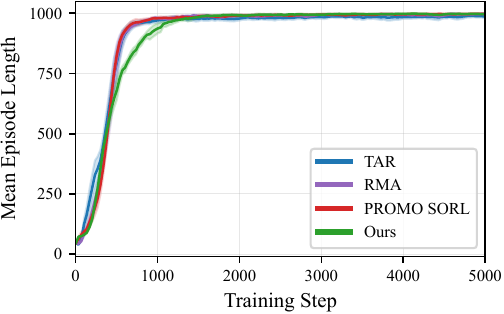}}\hfill
\subfloat[\acs{morl} baselines: episode length]{\includegraphics[width=0.32\linewidth]{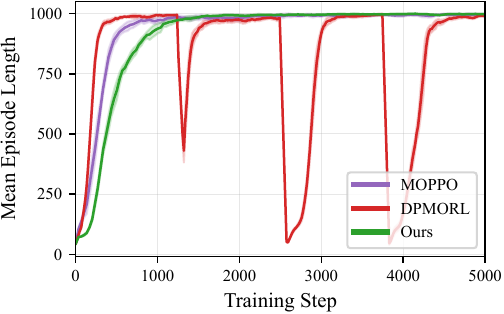}}\hfill
\subfloat[Specialists: episode length]{\includegraphics[width=0.32\linewidth]{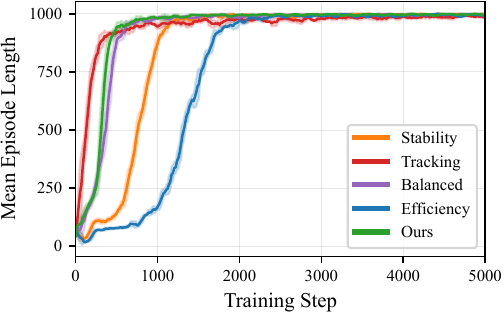}}
\caption{Training optimization over three seeds (mean, $\pm1$ s.d.). The top row reports mean reward and the bottom row reports mean episode length for locomotion baselines (a,d), \acs{morl} baselines (b,e), and independently trained specialists (c,f). \acs{promo} converges to a higher reward than the shared-scale locomotion and \acs{morl} references while matching the episode-length convergence of the balanced specialist, learning the preference space at roughly the optimization cost of a single controller. Specialist rewards are shown for completeness but use differently scaled scalar objectives, so episode length is the more informative cross-specialist comparison.}
\label{fig:train_curves}
\end{figure*}

\label{sec:baselines}
The training comparison evaluates mean reward and episode length over three independent seeds, with one-standard-deviation bands (Fig.~\ref{fig:train_curves}). These curves measure optimization under each method's native objective; deployable interface quality is assessed separately in Tables~\ref{tab:common_locomotion_eval} and~\ref{tab:anchor_preference_eval}.

\textbf{Locomotion baselines.} Against \ac{tar}, \ac{rma}, and the \ac{promo}-\ac{sorl} specialists, all methods rapidly acquire stable locomotion and reach the $1000$-step episode horizon. The single-objective controllers reach that horizon slightly earlier, but \ac{promo} continues improving after locomotion has stabilized, converging to $27.5$ reward with low seed variance, compared with $25.3$ for \ac{promo}-\ac{sorl}, $24.8$ for \ac{rma}, and $24.2$ for \ac{tar}. This indicates that the preference-conditioned formulation can preserve stable locomotion while amortizing multiple semantic behaviors into one policy.

\textbf{\acs{morl} baselines.} \ac{promo} and the single-policy \ac{moppo} baseline learn similarly early, but \ac{promo} continues to $27.5$ reward compared with $23.5$ for \ac{moppo}, with lower variance. \ac{dpmorl} shows four distinct optimization phases, one per separately trained policy, each transition resetting optimization. \ac{promo} instead learns the preference space in one continuous run with a single deployable actor.

\textbf{Amortization against specialists.} Specialist reward curves are not directly comparable because each policy optimizes a differently scaled scalar objective; episode length, with identical terminations, is the common axis. \ac{promo} reaches the horizon after $550$--$650$ steps, close to the balanced specialist ($600$--$700$) and tracking specialist ($400$--$500$), whereas the efficiency specialist converges much later ($1700$--$1900$) because effort minimization is a weak gait-discovery signal. Thus, \ac{promo} learns the four semantic behaviors at roughly the optimization cost of one balanced controller, instead of training and deploying four separate policies.

\subsection{Common Semantic Evaluation Against Baselines}
\label{sec:baseline_semantic_eval}
Training reward is insufficient to validate a preference-conditioned locomotion interface. We therefore separate balanced locomotion quality from preference-interface quality across requested behaviors.

\textit{Balanced locomotion quality.}
Table~\ref{tab:common_locomotion_eval} evaluates the balanced operating point. \ac{promo} records the highest mean tracking reward ($18.703$), compared with $16.225$ for the closest method, \ac{promo}-\ac{sorl}. It also has a lower mean fall rate than \ac{tar} and \ac{rma} ($7.4\%$ vs.\ $14.5\%$ and $18.2\%$) and the lowest mean \ac{cvar} tilt ($0.105$). The reward rankings remain objective-dependent: \ac{tar} has the highest mean aggregate stability reward, whereas \ac{rma} has the highest mean efficiency reward. The results also suggest a potential benefit of preference-conditioned training. By training a shared actor across tracking-, stability-, and efficiency-biased preferences, \ac{promo} is exposed to a broader behavioral repertoire that may benefit its balanced operating point.

\begin{table}[t]
\caption{Common balanced locomotion evaluation. Higher is better for semantic rewards; lower is better for fall and \acs{cvar} tilt. Fall values are percentages. Bold denotes the best mean in each column.}
\label{tab:common_locomotion_eval}
\centering
\footnotesize
\setlength{\tabcolsep}{4pt}
\resizebox{\columnwidth}{!}{%
\begin{tabular}{lccccc}
\toprule
Method & Tracking $\uparrow$ & Stability $\uparrow$ & Efficiency $\uparrow$ & Fall $\downarrow$ & \acs{cvar} tilt $\downarrow$ \\
\midrule
\ac{tar} & $14.921$ & $\mathbf{-2.218}$ & $-3.643$ & $14.5\%$ & $0.116$ \\
\ac{rma} & $13.742$ & $-2.920$ & $\mathbf{-3.423}$ & $18.2\%$ & $0.125$ \\
\ac{promo}-\ac{sorl} & $16.225$ & $-2.804$ & $-4.157$ & $11.0\%$ & $0.109$ \\
\textbf{\ac{promo} (ours)} & $\mathbf{18.703}$ & $-2.855$ & $-5.366$ & $\mathbf{7.4\%}$ & $\mathbf{0.105}$ \\
\bottomrule
\end{tabular}
}
\end{table}

\textit{Four-anchor preference interface.}
Table~\ref{tab:anchor_preference_eval} evaluates the balanced and three single-objective anchor requests. \ac{promo} records higher mean requested tracking and efficiency than the specialists ($20.099$ vs.\ $17.833$ and $-2.189$ vs.\ $-2.942$, respectively), while the stability specialist retains the higher mean stability score ($-1.916$ vs.\ $-2.245$). \ac{promo} also records a $4.2$-percentage-point lower aggregate fall mean relative to the specialist family ($8.4\%$ vs.\ $12.6\%$). The four-anchor results therefore show no clear numerical amortization penalty while demonstrating that one actor can express multiple semantic behaviors without matching every specialist extremum.

Compared with \ac{moppo}, \ac{promo} records a higher mean scalarized return ($3.964$ vs.\ $2.212$), an $11.0$-percentage-point lower mean fall rate, and higher mean four-anchor controllability ($0.662$ vs.\ $0.492$). \ac{dpmorl} and \ac{promo} have comparable mean requested efficiency ($-2.133$ vs.\ $-2.189$). \ac{dpmorl}, however, records a substantially higher mean fall rate of $26.6\%$. The comparison supports the intended deployment profile through the observed combination of objective specialization, robustness, preference response, and one-policy execution.

\begin{table}[t]
\caption{Four-anchor semantic evaluation against \acs{morl} and specialist baselines. Bold denotes the best mean in each column.}
\label{tab:anchor_preference_eval}
\centering
\scriptsize
\setlength{\tabcolsep}{3pt}
\resizebox{\columnwidth}{!}{%
\begin{tabular}{lcccccc}
\toprule
Method &
\makecell{Tracking \\$T \uparrow$} &
\makecell{Stability \\$S \uparrow$} &
\makecell{Efficiency \\$E \uparrow$} &
Fall $\downarrow$ &
\makecell{Scalarized\\return $\uparrow$} &
\makecell{4-anchor\\Ctrl. $\uparrow$} \\
\midrule
\ac{dpmorl} & $14.430$ & $-2.352$ & $\mathbf{-2.133}$ & $26.6\%$ & $2.986$ & $--$ \\
\ac{moppo} & $15.326$ & $-2.305$ & $-2.794$ & $19.4\%$ & $2.212$ & $0.492$ \\
Specialists & $17.833$ & $\mathbf{-1.916}$ & $-2.942$ & $12.6\%$ & $3.255$ & $--$ \\
\textbf{\ac{promo}} & $\mathbf{20.099}$ & $-2.245$ & $-2.189$ & $\mathbf{8.4\%}$ & $\mathbf{3.964}$ & $\mathbf{0.662}$ \\
\bottomrule
\end{tabular}
}
\end{table}

\subsection{Pareto Coverage and Preference Controllability}
\label{sec:promo_standalone_eval}
We evaluate \ac{promo} over $100$ preference vectors sampled uniformly over the simplex (Figure~\ref{fig:preference_simplex}), with each preference rolled out across $1024$ parallel environments for $20$~s under external disturbances. Appendix~\ref{app:evaluation_protocols}, Table~\ref{tab:controllability} summarizes the resulting Pareto quality and preference controllability.

\textit{Pareto quality.} The achieved semantic objective set provides a dense sampled Pareto-front approximation: $67$ of the $100$ preference-induced policies are non-dominated within the evaluated set under Eq.~\eqref{eq:pareto_dominance}. Applying $\varepsilon$-dominance pruning following~\cite{laumanns2002combining} (Appendix~\ref{sec:supp_figures}, Table~\ref{tab:supp_eps_dominance}) retains $37$ solutions at $\varepsilon=1\%$, indicating a densely populated trade-off surface rather than isolated anchors. Tracking and efficiency form the clearest conflict, while stability is partly complementary to both.

\begin{figure*}[t]
\centering
\subfloat[Tracking vs.\ efficiency front approximation]{\includegraphics[width=0.32\linewidth]{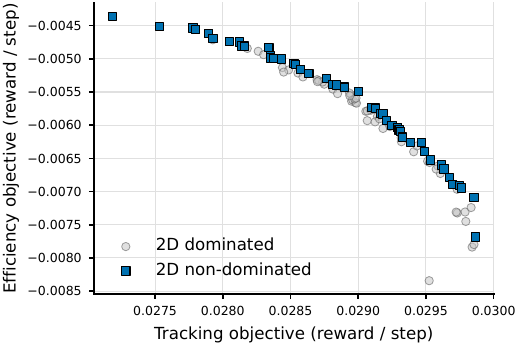}}\hfill
\subfloat[Tracking vs.\ stability front approximation]{\includegraphics[width=0.32\linewidth]{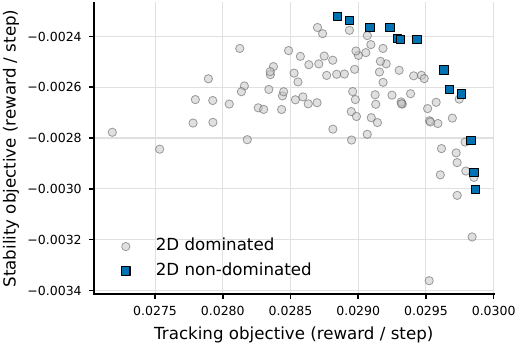}}\hfill
\subfloat[Stability vs.\ efficiency front approximation]{\includegraphics[width=0.32\linewidth]{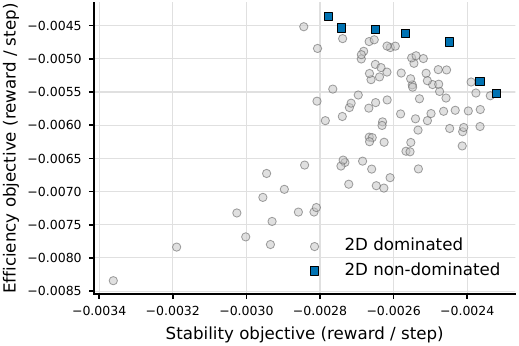}}\\[2pt]
\subfloat[Tracking response]{\includegraphics[width=0.32\linewidth]{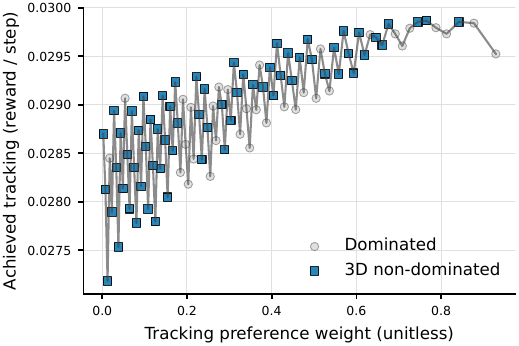}}\hfill
\subfloat[Stability response]{\includegraphics[width=0.32\linewidth]{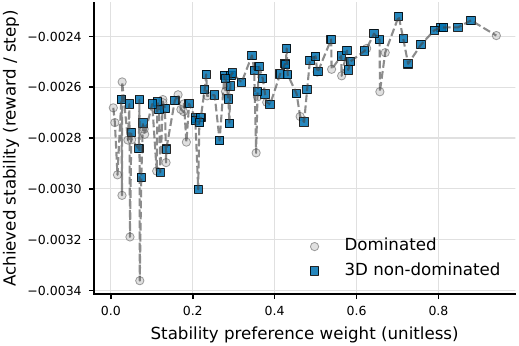}}\hfill
\subfloat[Efficiency response]{\includegraphics[width=0.32\linewidth]{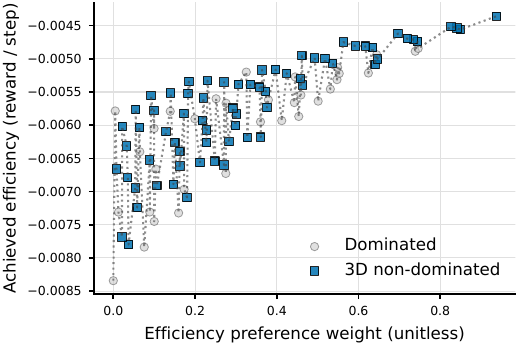}}
\caption{\textbf{Standalone simulation results for \ac{promo} policy evaluated at 100 sampled preference vectors.}
Each sampled preference produces one policy evaluation; panels (a)--(c) show different 2-D projections of these same 100 evaluations, with Pareto-nondominated and dominated samples distinguished visually. Tracking and efficiency form the clearest conflict, while stability and efficiency are largely complementary. Panels (d)--(f) show marginal per-objective preference-to-behavior mappings: each realized objective return rises near-monotonically with its corresponding preference weight, while the spread reflects variation in the other two simplex weights.
}
\label{fig:flagship}
\end{figure*}

\textit{Preference controllability.} The learned interface is strongly aligned with the requested preference: increasing a semantic weight generally corresponds to a higher realized return. The dense evaluation records $0.8434$ controllability and a $7.50\%$ mean fall rate, with failures concentrated near aggressive preference boundaries. Figure~\ref{fig:flagship} summarizes the Pareto geometry and preference-to-behavior mappings for our \ac{promo} policy.

The per-objective correlations $\rho_T,\rho_S,\rho_E$ show the largest response on the efficiency axis, while tracking and stability remain strongly aligned with their preference weights despite tighter coupling to gait dynamics.

Together, the dense results show that the four anchor behaviors are samples from a broader controllable interface rather than isolated operating points.

\section{Real-Robot Evaluation}
\label{sec:real_robot}
The hardware study tests whether the simulated semantic interface transfers to a physical robot (Q4). A single \ac{promo} policy trained in simulation is deployed on hardware without retraining or controller switching, and preference changes alone modulate measured tracking, stability, and efficiency behavior.

\subsection{Protocol}

All hardware experiments use a single deployed \ac{promo} checkpoint on a Unitree Go2~\cite{unitree_go2}. Experiments are conducted indoors with Vicon motion capture providing ground-truth position and velocity. We evaluate the four anchors in Fig.~\ref{fig:preference_simplex}: balanced $(\tfrac13,\tfrac13,\tfrac13)$, tracking-heavy $(0.8,0.1,0.1)$, stability-heavy $(0.1,0.8,0.1)$, and efficiency-heavy $(0.1,0.1,0.8)$, across three command regimes. The replay regime uses the same recorded trajectory for every preference and is the command-matched comparison. Fast and slow regimes use different route-speed profiles and are reported through aggregate outcomes. This gives $12$ conditions with five trials each. The policy runs at approximately 50\,Hz over a 200\,Hz low-level control loop; the observation stack, command interface, and network weights remain fixed. Table~\ref{tab:hardware} reports means, and Fig.~\ref{fig:hardware_semantic} shows run-to-run variability.

\subsection{Preference-Conditioned Behavior}

\begin{figure*}[t]
\centering
\subfloat[Tracking-heavy]{\includegraphics[width=0.33\textwidth]{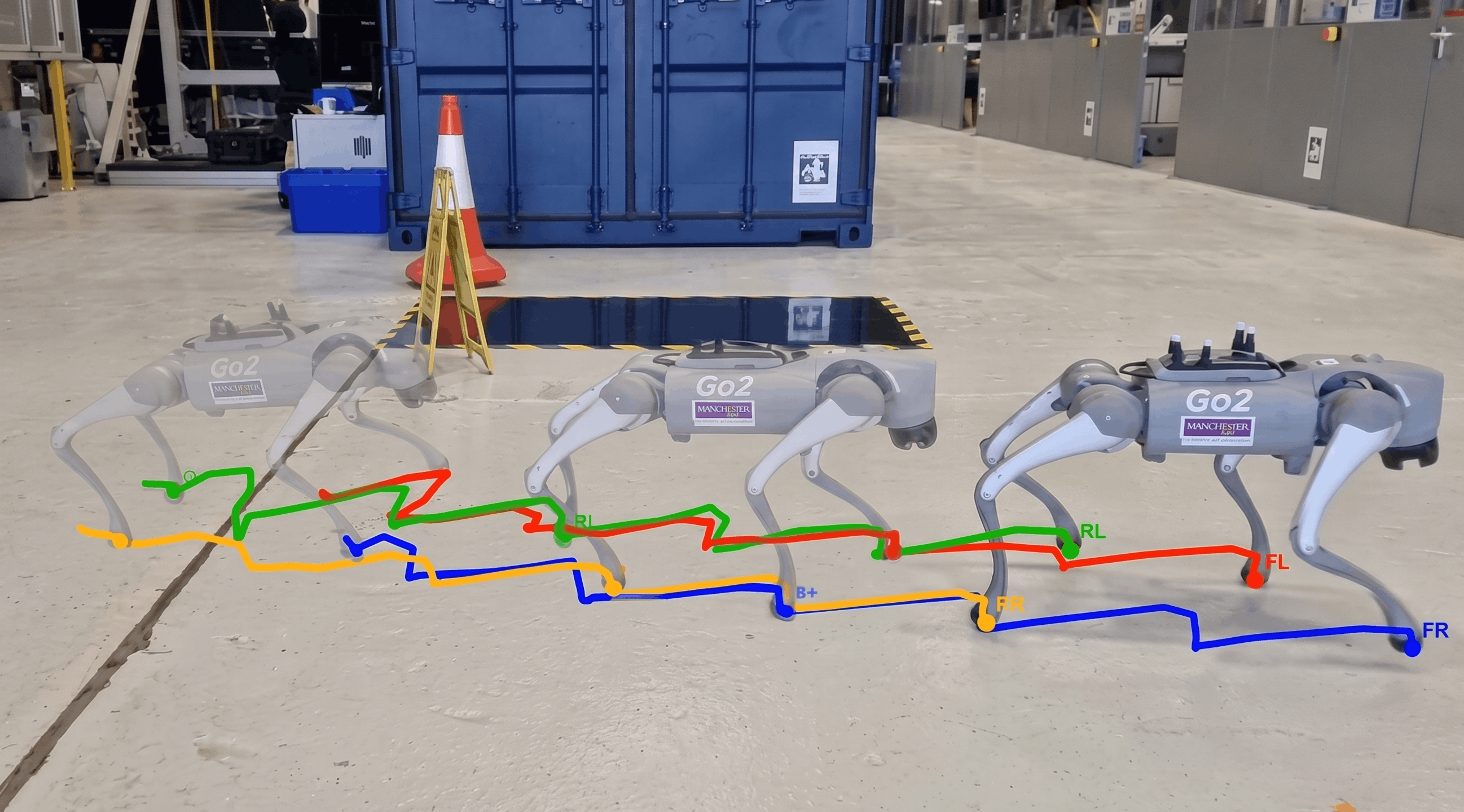}}\hfill
\subfloat[Stability-heavy]{\includegraphics[width=0.33\textwidth]{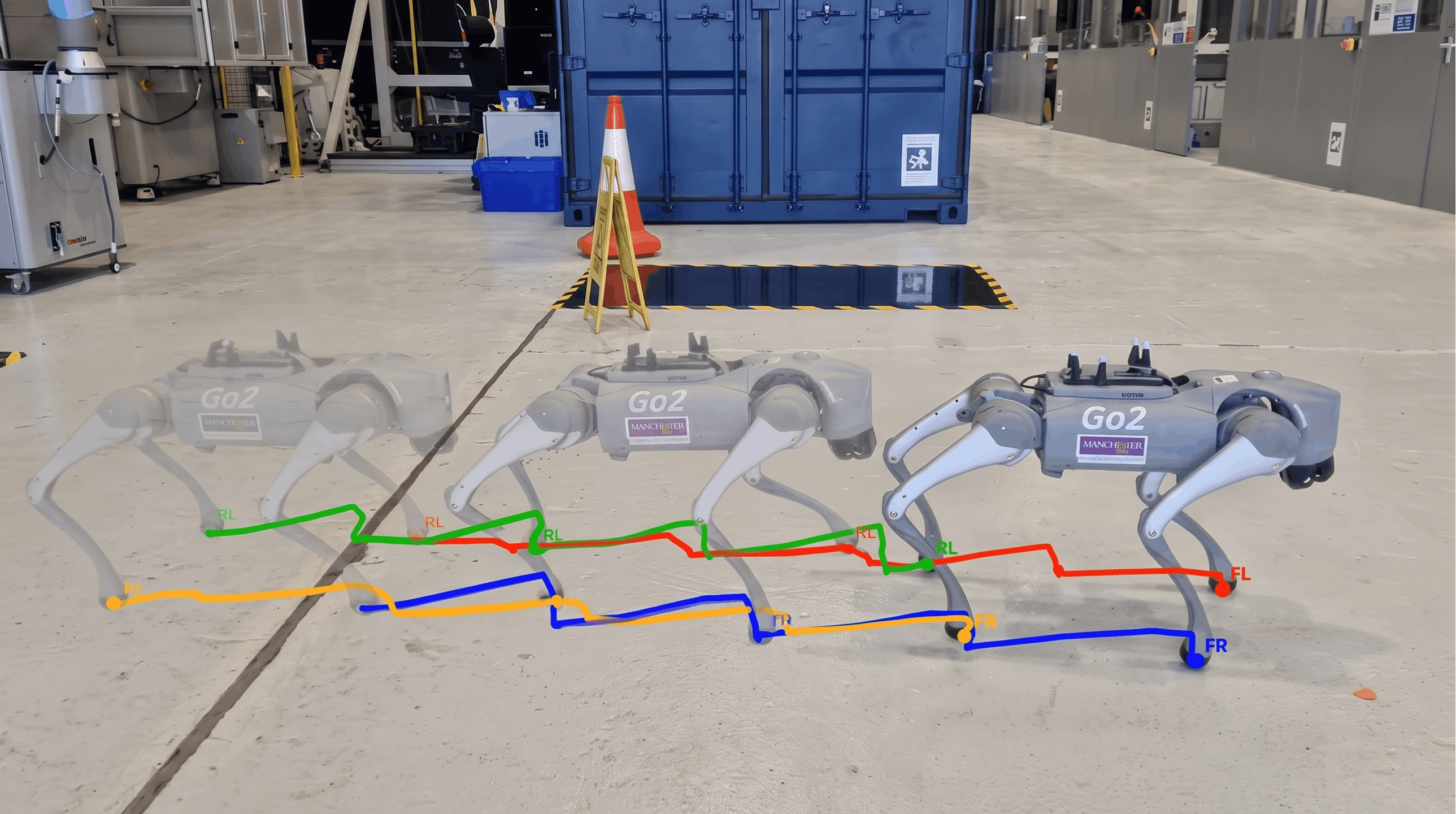}}\hfill
\subfloat[Efficiency-heavy]{\includegraphics[width=0.33\textwidth]{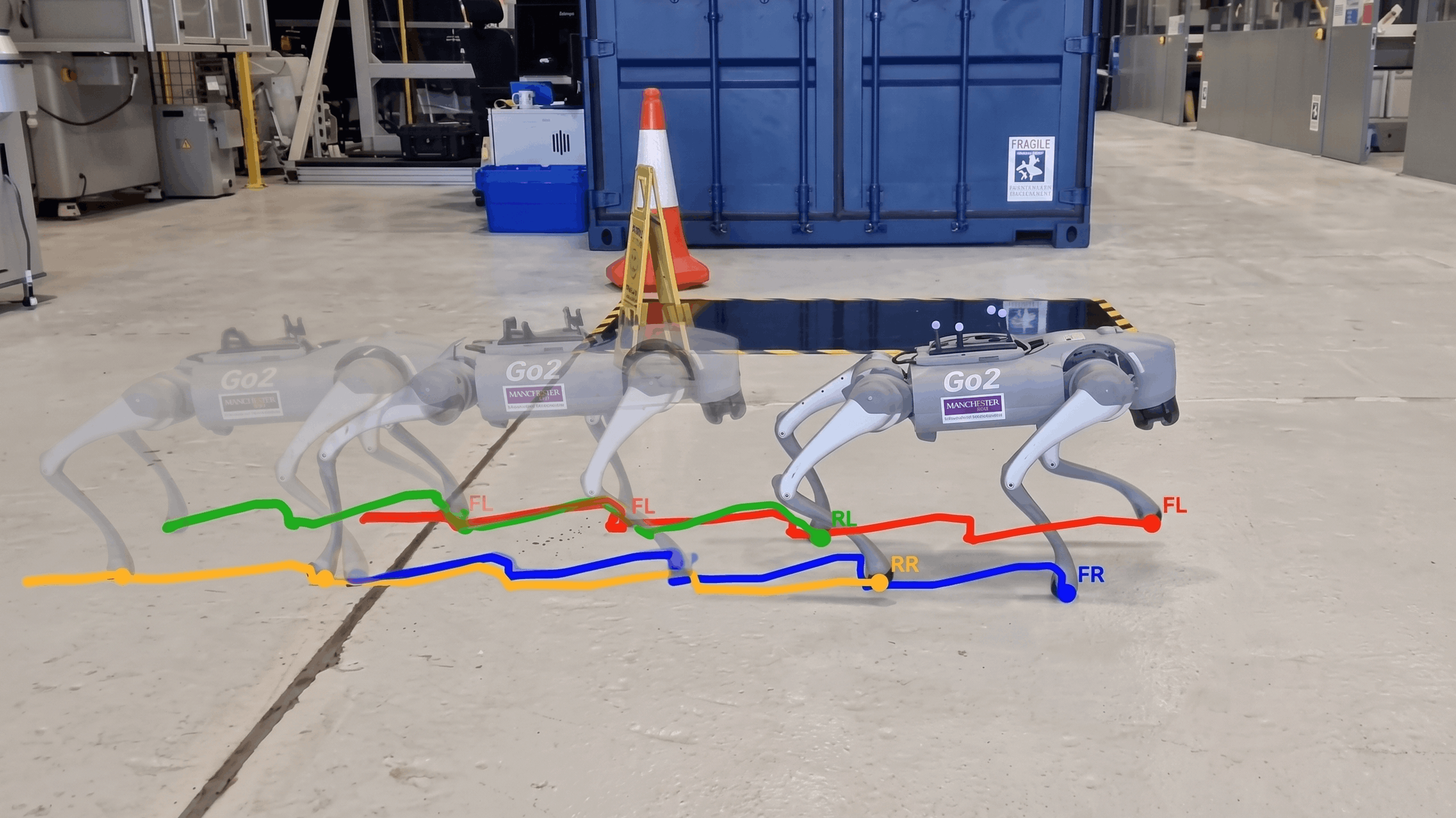}}
\caption{\textbf{Preference-induced gait adaptation on hardware.}
Chronophotographic overlays with 2-s tracked foot trajectories for tracking-, stability-, and efficiency-heavy preferences (left to right) using the same deployed \acs{promo} policy. Tracking produces larger foot excursions, efficiency produces compact low-clearance trajectories, and stability exhibits an intermediate gait with increased clearance in this sequence.}
\label{fig:hardware_chronophoto}
\end{figure*}

Figure~\ref{fig:hardware_chronophoto} shows the physical gait changes induced by preference, providing a glimpse of how the learned optimal actions adapt with changing preferences. The tracking-heavy preference produces larger foot excursions, whereas the efficiency-heavy preference yields compact, low-clearance trajectories. The stability-heavy behavior exhibits an intermediate gait with more pronounced foot clearance in this sequence. These differences emerge without prescribing gait parameters.

\begin{figure*}[t]
\centering
\includegraphics[width=0.90\textwidth]{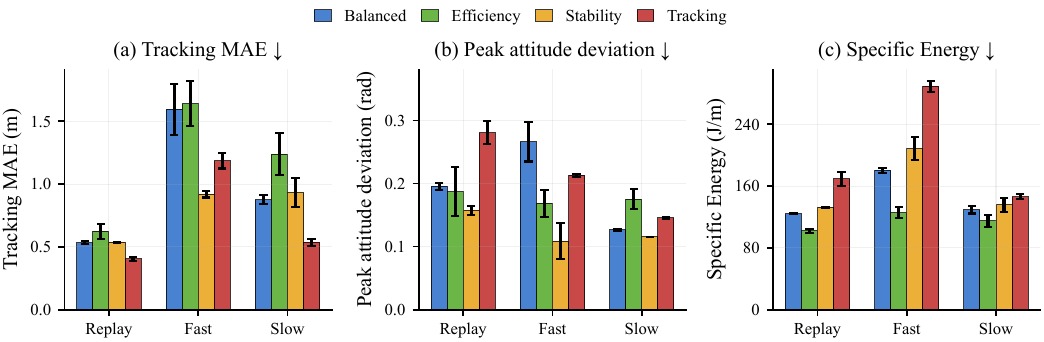}
\caption{\textbf{Preference-conditioned hardware behavior.}
Tracking \ac{mae}, peak attitude deviation, and specific energy for the same deployed \acs{promo} policy under four semantic preferences across replay, fast, and slow regimes. Bars show means over five hardware trials and error bars denote one standard deviation. Lower is better for all metrics. The corresponding preference generally shifts behavior toward its intended semantic objective, while cross-objective differences expose the resulting physical trade-offs.}
\label{fig:hardware_semantic}
\end{figure*}

Table~\ref{tab:hardware} and Fig.~\ref{fig:hardware_semantic} quantify these differences. The clearest effect is energetic: the efficiency preference records lower mean specific energy in all three regimes, by $17.9\%$, $30.4\%$, and $11.3\%$ relative to balanced in replay, fast, and slow, respectively. Action- and torque-rate RMS are also approximately $29\%$ lower on average, indicating a large, consistent shift toward less aggressive actuation.

The tracking preference records lower mean position-error \ac{mae} than balanced in all three regimes, by $24.2\%$, $25.5\%$, and $38.7\%$ in replay, fast, and slow, respectively. This tracking shift is accompanied by approximately $24$--$49\%$ higher action- and torque-rate RMS. In the fast regime, the stability preference also records a lower mean position \ac{mae} of $0.919$\,m, indicating that body regulation can improve trajectory progress under dynamic commands.

The stability preference reduces peak attitude deviation in every regime, by $19.4\%$, $59.0\%$, and $8.7\%$ relative to balanced in replay, fast, and slow, respectively. The effect is largest during dynamic locomotion and weaker near standstill. The balanced preference provides a compromise when no objective is prioritized, including a lower velocity RMSE in fast and slow operation. The slow-regime stability setting has high aggregate velocity RMSE ($2.203$\,m/s) despite low peak attitude deviation, showing that velocity regulation and body-attitude regulation capture different near-standstill failure modes.

These results establish the central deployment property of \ac{promo}: preference changes systematically modulate hardware behavior with a fixed policy. The objectives are coupled. For example, the tracking preference improves accumulated position error while slightly worsening velocity RMSE relative to balanced, consistent with more aggressive transient tracking that reduces long-horizon positional drift. We therefore report both measures rather than treating velocity RMSE alone as tracking performance.

\begin{table*}[t]
\caption{Real-robot evaluation of a single \acs{promo} policy under four semantic preferences and three command regimes on the Unitree Go2. Values are means over $n=5$ trials per condition; \textbf{bold} denotes the preferred value within each regime.}
\label{tab:hardware}
\centering
\footnotesize
\setlength{\tabcolsep}{5pt}
\renewcommand{\arraystretch}{1.05}

\begin{tabular}{lccccccc}
\toprule
Preference &
\makecell{Vel.\ RMSE\\{[m/s]}} &
\makecell{Pos.\ MAE\\{[m]}} &
\makecell{Energy\\{[J/m]}} &
\makecell{\shortstack{\acs{cot}\\{[-]}}} &
\makecell{Peak att.\\{[rad]}} &
\makecell{Act.\ rate\\{[rad/s]}} &
\makecell{Trq.\ rate\\{[N\,m/s]}} \\
\midrule
\multicolumn{8}{l}{\emph{Replay-round}}\\
Balanced & $0.444$ & $0.534$ & $124.6$ & $1.058$ & $0.196$ & $8.45$ & $240.7$ \\
Efficiency & $0.444$ & $0.624$ & $\mathbf{102.3}$ & $\mathbf{0.869}$ & $0.187$ & $\mathbf{5.42}$ & $\mathbf{153.4}$ \\
Stability & $\mathbf{0.428}$ & $0.538$ & $132.4$ & $1.125$ & $\mathbf{0.158}$ & $9.26$ & $263.0$ \\
Tracking & $0.493$ & $\mathbf{0.405}$ & $169.6$ & $1.441$ & $0.281$ & $11.98$ & $342.9$ \\
\midrule
\multicolumn{8}{l}{\emph{Move-fast}}\\
Balanced & $\mathbf{0.437}$ & $1.594$ & $180.5$ & $1.533$ & $0.266$ & $9.72$ & $276.3$ \\
Efficiency & $0.737$ & $1.641$ & $\mathbf{125.7}$ & $\mathbf{1.068}$ & $0.168$ & $\mathbf{6.22}$ & $\mathbf{175.8}$ \\
Stability & $0.841$ & $\mathbf{0.919}$ & $208.7$ & $1.610$ & $\mathbf{0.109}$ & $9.89$ & $280.3$ \\
Tracking & $0.753$ & $1.187$ & $289.3$ & $2.458$ & $0.213$ & $14.09$ & $411.1$ \\
\midrule
\multicolumn{8}{l}{\emph{Move-slow}}\\
Balanced & $\mathbf{0.332}$ & $0.876$ & $129.6$ & $1.101$ & $0.127$ & $6.50$ & $182.7$ \\
Efficiency & $0.770$ & $1.239$ & $\mathbf{114.9}$ & $\mathbf{0.976}$ & $0.175$ & $\mathbf{5.53}$ & $\mathbf{156.6}$ \\
Stability & $2.203$ & $0.931$ & $135.8$ & $1.154$ & $\mathbf{0.116}$ & $5.96$ & $165.8$ \\
Tracking & $0.438$ & $\mathbf{0.537}$ & $146.5$ & $1.245$ & $0.146$ & $8.11$ & $224.2$ \\
\bottomrule
\end{tabular}
\end{table*}

\begin{figure*}[t]
\centering
\includegraphics[width=\textwidth]{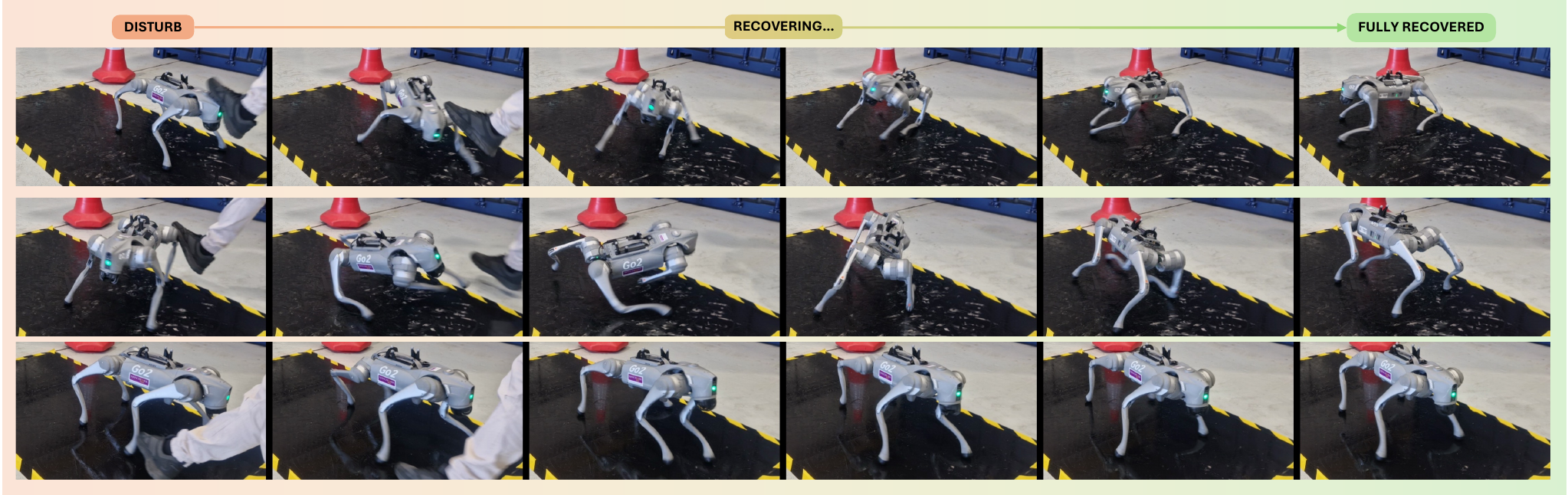}
\caption{\textbf{Disturbance recovery on a low-traction surface.}
Representative sequences using the balanced preference on a low-traction surface. Top is whole-body recovery following a disturbance at the head. Middle is recovery from a rear-left hip disturbance inducing approximately $230^\circ$ of yaw rotation. Bottom is recovery from direct disturbance of the front-right leg, requiring adjustment of the support configuration.}
\label{fig:hardware_slip_recovery}
\end{figure*}

Figure~\ref{fig:hardware_slip_recovery} examines low-traction recovery using the balanced preference. Despite substantial slip and displacement, the policy restores viable locomotion after whole-body, rotational, and direct limb disturbances; the most severe rotational example produces approximately $230^\circ$ of yaw rotation before recovery. Because disturbance forces were not instrumented, these sequences are qualitative stress tests rather than force-normalized robustness measurements. Complete recovery trajectories are provided in the accompanying videos.

\section{Discussion}
\label{sec:discussion}

This section interprets \ac{promo} as a deployment interface, distinguishes behavioral control from offline Pareto optimization, and discusses the limitations of the present study.

\subsection{Semantic Preferences as a Deployment Interface}

\ac{promo} treats preference-conditioned \ac{morl} as a runtime locomotion interface for reward non-stationarity. The non-stationary quantity is the scalar reward specification induced by $w_t$, not the transition dynamics. This framing targets a practical setting in which the relevant objective classes are known, but their desired trade-off changes during operation. The hardware results show that this trade-off can be exposed as a semantic input to one real robot policy rather than resolved offline into a fixed controller.

The semantic objective design is more than reward engineering. The policy remains a neural controller, but the operator-facing cause of a behavioral change is expressed through meaningful objective weights rather than an opaque scalar reward or latent gait code. This is not a causal explanation of every joint command; it is a transparent interface at the level where deployment decisions are made, allowing behavior to be inspected against the intended tracking, stability, and efficiency emphasis.

\subsection{From Pareto Optimization to Deployable Control}
The results and ablations in Appendix~\ref{app:analysis} support two design conclusions. First, model selection for a deployed preference interface should balance utility, lower-tail rollout outcomes, controllability, and fall rate rather than optimize any single aggregate metric. Second, locomotion priors should remain fixed auxiliary terms: removing them, embedding them in the semantic objectives, or exposing them as a fourth objective can improve an aggregate metric while degrading controllability or reliability.

The baseline evaluation makes the same distinction at the method-comparison level. Individual objective extrema and deployment-interface quality are different criteria: several baselines retain an advantage in one semantic score, whereas \ac{promo} provides a stronger combination of objective specialization, lower failures, preference response, and one-policy deployment.

The results also suggest a second role for preference coverage during training. Sampling a broad semantic simplex appears to diversify the behaviors used to solve locomotion, so the shared policy is not simply interpolating among independently learned fixed-objective solutions. This may explain why the balanced \ac{promo} preference outperforms fixed-objective baselines on tracking, fall rate, and tail attitude risk. The evidence supports this as a plausible mechanism; a causal ablation of preference-coverage width remains future work.

The ablations show that reward-optimization quality and deployable-control quality are separable. Variants that match or exceed the baseline in training reward or range-normalized-objective \ac{hv} in the three semantic dimensions can underperform in controllability and falls, including variants that equalize objective magnitudes and suppress the tracking signal that drives gait acquisition (Appendix~\ref{app:rewards_preferences}). Robotic \ac{morl} should therefore evaluate Pareto quality together with whether preferences remain meaningful, controllable, and robust in the deployed system.

\subsection{Limitations}
\label{sec:limitations}

\ac{promo} targets operator-induced objective non-stationarity rather than non-stationary \acp{mdp} with changing transition dynamics, hidden regime shifts, or online adaptation. Linear scalarization focuses the policy on supported Pareto solutions and does not recover unsupported non-convex regions of the Pareto front. The critic estimates expected semantic returns, so lower-tail quantities such as \ac{cvar}$_{10}$ are rollout diagnostics rather than directly optimized critic objectives. The hardware study uses five trials per condition and four anchor preferences, demonstrating preference modulation rather than dense hardware controllability or statistical significance. The shared baseline evaluation likewise uses representative balanced and anchor settings; dense 100-preference baseline comparisons and ablations over training preference coverage remain future work.

\section{Conclusion}
\label{sec:conclusion}

We presented \ac{promo}, a semantic preference-conditioned \ac{morl} framework that exposes tracking, stability, and efficiency as runtime objectives for a single quadruped locomotion policy. By separating these semantic objectives from fixed locomotion priors, \ac{promo} provides an interpretable preference input for deployment. In simulation, the policy achieves $0.8434$ dense preference--objective correlation with a $7.50\%$ mean fall rate over 100 sampled preferences, and the baseline comparisons show that individual objective extrema and deployment-interface quality are distinct criteria.

On hardware, the same Unitree Go2 policy modulates energy use, trajectory error, and body attitude through preference changes alone, reducing specific energy, position error, and peak attitude deviation by up to $30.4\%$, $38.7\%$, and $59.0\%$ relative to the balanced setting. These results support semantic \ac{morl} as a deployment mechanism in which the robot exposes the objective trade-off itself as a runtime control input, rather than only producing an offline Pareto set after training.



\bibliographystyle{IEEEtran}
\bibliography{./references}

\appendices
\makeatletter
\@addtoreset{figure}{section}
\@addtoreset{table}{section}
\makeatother
\renewcommand{\thefigure}{\Alph{section}.\arabic{figure}}
\renewcommand{\thetable}{\Alph{section}.\arabic{table}}

\clearpage
\section{Reward and Preference Specification}
\label{app:rewards_preferences}

This appendix defines the reward quantities used by \ac{promo}. Semantic terms are grouped into tracking, stability, and efficiency objectives; fixed locomotion priors remain auxiliary terms outside the deployment preference vector.

\subsection{Semantic Reward Decomposition}

Tracking measures command following, stability measures body regulation, and efficiency measures energetic and control effort. Fixed locomotion priors encode embodiment-level regularity needed for viable legged motion. Table~\ref{tab:reward_terms_app_c} lists the active reward terms and environment coefficients.

\begin{table*}[b]
\caption{Semantic objectives and fixed locomotion priors used by \acs{promo}.}
\label{tab:reward_terms_app_c}
\centering
\scriptsize
\setlength{\tabcolsep}{3pt}
\begin{tabular}{p{0.28\textwidth}p{0.13\textwidth}p{0.12\textwidth}p{0.37\textwidth}}
\toprule
Term & Group & Coefficient & Physical interpretation \\
\midrule
\texttt{track\_lin\_vel\_xy\_exp} & Tracking & $1.5$ &
Rewards planar velocity tracking with kernel standard deviation $0.5$ \\
\texttt{track\_ang\_vel\_z\_exp} & Tracking & $0.75$ &
Rewards yaw-rate tracking with kernel standard deviation $0.5$ \\
\midrule
\texttt{lin\_vel\_z\_l2} & Stability & $-2.0$ &
Penalizes vertical base motion \\
\texttt{ang\_vel\_xy\_l2} & Stability & $-0.05$ &
Penalizes roll and pitch angular velocity \\
\texttt{flat\_orientation\_l2} & Stability & $-2.5$ &
Penalizes body tilt away from upright posture \\
\midrule
\texttt{joint\_torques\_l2} & Efficiency & $-2.0\times10^{-4}$ &
Penalizes squared applied torque \\
\texttt{joint\_acc\_l2} & Efficiency & $-2.5\times10^{-7}$ &
Penalizes squared joint acceleration \\
\texttt{action\_rate\_l2} & Efficiency & $-0.01$ &
Penalizes rapid changes in action commands \\
\midrule
\texttt{feet\_air\_time} & Fixed loco. prior & $0.25$ &
Promotes viable stepping with minimum swing time $0.5$\,s \\
\texttt{foot\_contact\_balance\_penalty} & Fixed loco. prior & $-0.5$ &
Penalizes imbalanced foot contact over a $1.0$\,s window \\
\texttt{joint\_pos\_penalty} & Fixed loco. prior & $-0.1$ &
Regularizes joint posture, with stronger stand-still weighting \\
\bottomrule
\end{tabular}
\end{table*}

The scalar semantic reward used for actor optimization is
\begin{equation}
r_t^{\mathrm{sem}}(w_t) =
w_{t,\mathrm{track}}r_t^{\mathrm{track}}
+w_{t,\mathrm{stab}}r_t^{\mathrm{stab}}
+w_{t,\mathrm{eff}}r_t^{\mathrm{eff}},
\end{equation}
with $w\in\Delta^2$. The fixed locomotion priors form $r_t^{\mathrm{prior}}$ and are added to the actor-side utility, but they are not entries of $w_t$ and do not define critic heads. All reported \ac{promo} experiments use $\lambda_{\mathrm{prior}}=1$ in Eq.~\eqref{eq:training_utility_main}, so the actor reward is
\begin{equation}
r_t^{\mathrm{actor}}
=
r_t^{\mathrm{sem}}(w_t)
+
r_t^{\mathrm{prior}}.
\end{equation}

\subsection{Per-Step Objective Definitions}

Let $c_t=(c^x_t,c^y_t,c^\psi_t)$ be the commanded planar velocity and yaw rate, $v^b_t$ and $\omega^b_t$ the base linear and angular velocity in the body frame, $g^b_t$ the projected gravity vector, $q_t$, $\dot q_t$, $\ddot q_t$, and $\tau_t$ the joint state and applied torque, and $a_t$ the policy action. The raw quantities grouped in Table~\ref{tab:reward_terms_app_c} are defined below.

The tracking objective uses exponential kernels,
\begin{align}
r^{\mathrm{lin}}_t
&=
\exp\left(
-\frac{(c^x_t-v^{b,x}_t)^2+(c^y_t-v^{b,y}_t)^2}{\sigma_v^2}
\right),\\
r^{\mathrm{yaw}}_t
&=
\exp\left(
-\frac{(c^\psi_t-\omega^{b,z}_t)^2}{\sigma_\psi^2}
\right),
\end{align}
with $\sigma_v=\sigma_\psi=0.5$.

The stability penalties are
\begin{align}
p^{z\text{-vel}}_t &= (v^{b,z}_t)^2,\\
p^{xy\text{-ang}}_t &= (\omega^{b,x}_t)^2+(\omega^{b,y}_t)^2,\\
p^{\mathrm{orient}}_t &= (g^{b,x}_t)^2+(g^{b,y}_t)^2.
\end{align}
The efficiency penalties are
\begin{align}
p^\tau_t &= \sum_j (\tau^j_t)^2,\\
p^{\ddot q}_t &= \sum_j (\ddot q^j_t)^2,\\
p^{\Delta a}_t &= \sum_j (a^j_t-a^j_{t-1})^2.
\end{align}
These terms are implemented as sums over the configured joints or action dimensions, not as dimension-normalized means.

The semantic objectives used in $r_t^{\mathrm{sem}}$ are therefore
\begin{align}
r_t^{\mathrm{track}} &=
1.5\,r^{\mathrm{lin}}_t
+0.75\,r^{\mathrm{yaw}}_t,\\
r_t^{\mathrm{stab}} &=
-2.0\,p^{z\text{-vel}}_t
-0.05\,p^{xy\text{-ang}}_t
-2.5\,p^{\mathrm{orient}}_t,\\
r_t^{\mathrm{eff}} &=
-2.0{\times}10^{-4}\,p^\tau_t
-2.5{\times}10^{-7}\,p^{\ddot q}_t
-0.01\,p^{\Delta a}_t.
\end{align}

Fixed locomotion priors are optimized during training but remain independent of the semantic preference space. The feet air-time prior rewards sufficiently long swing phases upon first contact,
\begin{equation}
r^{\mathrm{air}}_t
=
\mathbb{1}\!\left[\|c^{xy}_t\|_2>0.1\right]
\sum_{f\in\mathcal{F}}
\left(T^f_{\mathrm{air},t}-T_{\min}\right)
\mathbb{1}^{f}_{\mathrm{first\ contact},t},
\end{equation}
where $T_{\min}=0.5$\,s and the reward is active only under a non-negligible planar command.

The contact-balance prior discourages persistent imbalance in foot usage over a rolling window. For each foot $f$, the contact fraction is
\begin{equation}
\phi^f_t
=
\frac{1}{N_t}
\sum_{\tau\in\mathcal{W}_t}
\mathbb{1}\!\left[\|F^f_\tau\|_2>1.0\,\mathrm{N}\right],
\end{equation}
and define
\begin{equation}
p^{\mathrm{contact}}_t
=
\operatorname{Var}_{f\in\mathcal{F}}(\phi^f_t)
+
\frac{1}{|\mathcal{F}|}
\sum_{f\in\mathcal{F}}
\mathbb{1}\!\left[\phi^f_t<\phi_{\min}\right].
\end{equation}
The first term penalizes unequal contact utilization across feet, while the second penalizes feet that are rarely used. We use a $1.0$\,s window and $\phi_{\min}=0.05$.

The joint-position prior regularizes the robot toward its nominal posture,
\begin{equation}
p^{\mathrm{jpos}}_t
=
\eta_t\left\|q_t-q^{0}\right\|_2,
\end{equation}
where $\eta_t=1$ during locomotion and $\eta_t=5$ when standing still, imposing stronger posture regularization near stationary operation. The recovered configuration defines locomotion as active when either the command norm exceeds $0.1$ or the planar body-speed norm exceeds $0.5$\,m/s.

\subsection{Preference Space and Sampling}

Training rollouts use fixed sampled preferences. The default training distribution samples $x\sim\mathrm{Dirichlet}(1,1,1)$ and maps it affinely to the interior of the semantic simplex,
\begin{equation}
w_{t,i}=w_{\min}+(1-Kw_{\min})x_i,
\qquad
w_{\min}=0.1,\quad K=3.
\end{equation}
Thus $w_{t,i}\ge0.1$ and $\sum_i w_{t,i}=1$ automatically; no clipping or projection is applied after sampling. Because the Dirichlet density is uniform on the full simplex and the affine map has constant Jacobian, this procedure uniformly samples the truncated simplex with minimum component $0.1$. The minimum prevents any semantic objective from disappearing during training; in particular, zero tracking weight can remove the command-following signal needed to learn locomotion. The sampled preference is held fixed for the rollout used to estimate critic targets and actor advantages. Table~\ref{tab:preference_protocol_app_c} lists the training and evaluation preference configurations.

\begin{table*}[t]
\caption{Preference configurations used for training and evaluation.}
\label{tab:preference_protocol_app_c}
\centering
\scriptsize
\setlength{\tabcolsep}{3pt}
\renewcommand{\arraystretch}{1.0}
\begin{tabular}{p{0.18\linewidth}p{0.32\linewidth}p{0.40\linewidth}}
\toprule
Protocol & Preference specification & Purpose \\
\midrule
Training distribution & $w_i=0.1+0.7x_i$, $x\sim\mathrm{Dirichlet}(1,1,1)$ & Uniform truncated-simplex coverage without zeroed objectives \\
Balanced & $(1/3,1/3,1/3)$ & Nominal semantic trade-off \\
Tracking-heavy & $(0.8,0.1,0.1)$ & Tracking-focused anchor \\
Stability-heavy & $(0.1,0.8,0.1)$ & Stability-focused anchor \\
Efficiency-heavy & $(0.1,0.1,0.8)$ & Efficiency-focused anchor \\
Two-objective mixtures & $(0.45,0.45,0.1)$, $(0.45,0.1,0.45)$, $(0.1,0.45,0.45)$ & Representative simplex edges \\
Dense 100-preference evaluation & Stored set of $100$ full-simplex preferences & Pareto and controllability evaluation beyond truncated training support \\
\bottomrule
\end{tabular}
\end{table*}

\subsection{Objective Scaling and Semantic Interpretation}
\label{app:normalization}

\ac{promo} retains the nominal reward scales of the locomotion environment instead of normalizing the semantic objectives to equal numerical magnitudes before scalarization. The balanced anchor $(\tfrac13,\tfrac13,\tfrac13)$ therefore corresponds to the nominal scalar reward already tuned for viable locomotion, and other preferences act as relative semantic modifiers around that operating point.

The objectives do not play symmetric roles during gait acquisition. Tracking supplies the main locomotion-driving signal, while stability and efficiency act mainly as regularizing pressures. Consistent with this interpretation, scaling stability and efficiency by $2$--$5\times$ relative to tracking degraded locomotion quality and preference-conditioned behavior.

Retaining nominal scales lets the balanced preference recover the default trade-off while preference changes shift emphasis toward the named objectives. Consequently, absolute scalar utility values are tied to these nominal scales and should not be interpreted as scale-normalized objective importance.


\section{Architecture and Training Configuration}
\label{app:objective_algorithm}
\label{app:training_sim_config}

This appendix provides the reproducibility configuration for \ac{promo}: network and control settings, domain randomization, rollout collection, and optimizer constants.

\subsection{Network, Simulation, and Control Configuration}

Table~\ref{tab:setup} defines the architecture, simulator, and control stack used by \ac{promo}. Table~\ref{tab:domain_randomization} defines the randomized training environment used for robustness and zero-shot transfer.

\begin{table}[t]
\caption{\acs{promo} network, simulation, and control configuration.}
\label{tab:setup}
\centering
\footnotesize
\setlength{\tabcolsep}{4pt}
\begin{tabular}{p{0.39\linewidth}p{0.51\linewidth}}
\toprule
Item & Value \\
\midrule
Robot & Unitree Go2, $12$-\ac{dof} quadruped \\
Simulator & Isaac Sim / IsaacLab \\
Actor hidden sizes & $[512,256,128]$ \\
Critic shared trunk & $[512,256,128]$ \\
Semantic critic heads & $K=3$ scalar heads \\
Privileged encoder & MLP over privileged state and preference \\
History encoder & MLP over $10$-step history and preference \\
Privileged latent dimension & $45$ \\
Actor observation history & $10$ control steps \\
Velocity estimator input & $z_t^H$ and $o_{t-4:t}$ \\
Velocity estimator target & Three-dimensional base linear velocity $v_t^b$ \\
Action dimension & $12$ joint-position targets \\
Command ranges & $v_x,v_y\!\in\![-1,1]$\,m/s, $\dot\psi\!\in\![-1,1]$\,rad/s \\
Sim timestep & $5$\,ms ($200$\,Hz); control decimation $4$ \\
Policy / control rate & $50$\,Hz \\
Episode length & $20$\,s ($1000$ policy steps) \\
Terrain & Flat \\
\bottomrule
\end{tabular}
\end{table}

\begin{table}[t]
\caption{Domain randomization and perturbation ranges used during training.}
\label{tab:domain_randomization}
\centering
\footnotesize
\setlength{\tabcolsep}{4pt}
\begin{tabular}{p{0.42\linewidth}p{0.48\linewidth}}
\toprule
Parameter & Range / setting \\
\midrule
Static and dynamic friction & $[0.6,\,1.2]$ \\
Restitution coefficient & $[0,\,1]$ \\
Base mass offset & $[0,\,2]$~kg \\
Force perturbation & $\pm10$~N \\
Torque perturbation & $\pm5$~N\,m \\
Velocity perturbation & $\pm2$~m/s every $2$--$4$~s \\
Initial linear velocity & $\pm0.5$~m/s \\
Initial angular velocity & $\pm0.5$~rad/s \\
\bottomrule
\end{tabular}
\end{table}

\subsection{Rollout and Optimization Configuration}

The semantic critic uses objective-wise TD targets. The actor advantage uses the scalarized semantic critic as its baseline, while the fixed prior reward enters the \ac{gae} residual directly without a separate prior value head. Advantages are normalized per batch before the \ac{ppo} update. Table~\ref{tab:optimization_constants} lists the rollout and optimizer constants.

\begin{table}[!t]
\centering
\caption{Training and optimization configuration.}
\label{tab:optimization_constants}
\footnotesize
\setlength{\tabcolsep}{4pt}
\begin{tabular}{p{0.48\linewidth}p{0.40\linewidth}}
\toprule
Parameter & Value \\
\midrule
Parallel environments & $4096$ \\
Rollout horizon & $50$ policy steps per env \\
Discount factor $\gamma$ & $0.99$ \\
\ac{gae} parameter $\lambda_{\mathrm{GAE}}$ & $0.95$ \\
\ac{ppo} clipping $\epsilon$ & $0.12$ \\
Entropy coefficient & $0.005$ \\
Learning rate & Adaptive, min: $5{\times}10^{-6}$, max: $2{\times}10^{-4}$ \\
Target KL divergence & $0.006$ \\
Gradient-norm clipping & $0.5$ \\
Optimization epochs & $3$ \\
Minibatches & $4$ \\
Advantage normalization & Per batch \\
Optimizer & Adam \\
Latent-alignment coefficient $\lambda_{\mathrm{align}}$ & $1.0$ \\
Diversity coefficient $\lambda_{\mathrm{div}}$ & $10^{-5}$ \\
\bottomrule
\end{tabular}
\end{table}


\section{Evaluation Protocol}
\label{app:evaluation_protocols}
This appendix defines the simulation and hardware execution protocols.
\subsection{Simulation Evaluation Protocol}

\begin{figure}[t]
\centering
\includegraphics[width=0.8\linewidth]{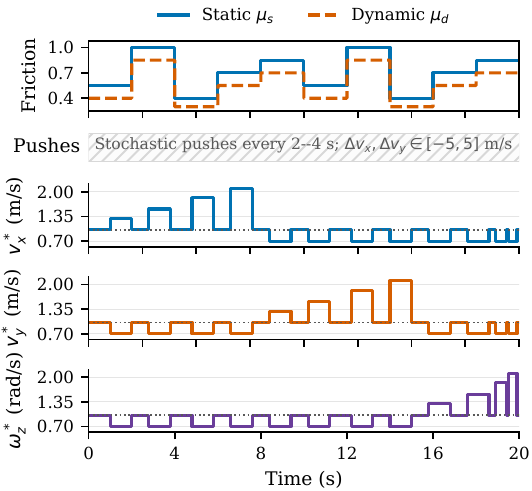}
\caption{Simulation evaluation protocol. Representative preferences reuse the same command-burst and friction/contact schedule, controlling for test-sequence variation when comparing preference-conditioned behavior.}
\label{fig:eval_protocol}
\end{figure}

Simulation evaluation uses two modes. The representative-preference protocol evaluates the seven anchor and mixed preferences in Table~\ref{tab:preference_protocol_app_c}, using $1024$ environments per preference over $20$\,s episodes. To isolate preference-induced differences, every representative evaluation uses the same disturbance script: the contact model alternates between predefined friction pairs every $2$~s while commanded forward, lateral, and yaw velocities follow a fixed burst sequence.

\label{sec:supp_figures}
The dense protocol evaluates $100$ sampled preferences, again with $1024$ environments per preference over $20$\,s episodes under standardized conditions. Each preference is executed independently, producing preference-performance maps for Pareto coverage, controllability, expected utility, lower-tail utility, and fall rate. Table~\ref{tab:supp_eps_dominance} gives the corresponding $\varepsilon$-dominance pruning. The drop from $67$ policies that are non-dominated within the sampled set under exact Pareto dominance to $37$ at $\varepsilon=1\%$ indicates a dense approximation of the trade-off surface rather than isolated anchors. The retained percentage is computed relative to the original exact non-dominated set.

\begin{table}[t]
\caption{Key interface metrics for \acs{promo} over $100$ preferences.}
\label{tab:controllability}
\centering
\footnotesize
\setlength{\tabcolsep}{5pt}
\begin{tabular}{lc}
\toprule
Metric & Value \\
\midrule
Mean utility per step & $0.0149$ \\
\ac{cvar}$_{10}$ utility & $0.0120$ \\
Fall rate & $7.50\%$ \\
Preference--objective correlation & $0.8434$ \\
\quad tracking $\rho_T$ & $0.7748$ \\
\quad stability $\rho_S$ & $0.8289$ \\
\quad efficiency $\rho_E$ & $0.9266$ \\
Range-normalized-objective \ac{hv} & $0.8266$ \\
Non-dominated samples (exact) & $(67)/100$ \\
\bottomrule
\end{tabular}
\end{table}

\begin{table}[t]
\caption{$\varepsilon$-dominance pruning of the 100-preference evaluation.}
\label{tab:supp_eps_dominance}
\centering
\footnotesize
\setlength{\tabcolsep}{5pt}
\begin{tabular}{lccccc}
\toprule
$\varepsilon$ threshold & $0$ & $0.005$ & $0.01$ & $0.02$ & $0.03$ \\
\midrule
Non-dominated solutions & $67$ & $50$ & $37$ & $25$ & $14$ \\
Non-dominated set retained (\%) & $100$ & $74.6$ & $55.2$ & $37.3$ & $20.9$ \\
\bottomrule
\end{tabular}
\end{table}

\begin{figure}[t]
\centering
\includegraphics[width=0.78\linewidth]{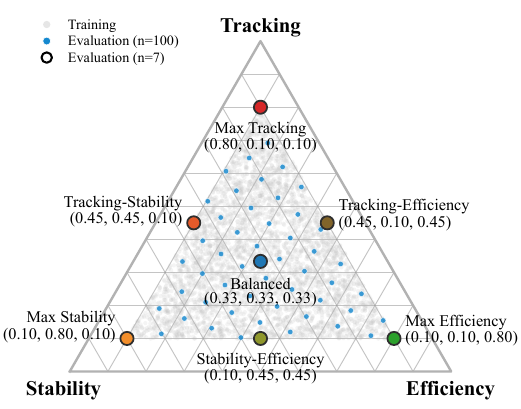}
\caption{Semantic preference simplex for \acs{promo}. The simplex defines the deployment-facing trade-off among tracking, stability, and efficiency; the marked anchors are used for representative simulation and hardware evaluations, while the dense evaluation samples the interior.}
\label{fig:preference_simplex}
\end{figure}

Figures~\ref{fig:eval_protocol} and~\ref{fig:preference_simplex} summarize the scripted evaluation protocol and semantic preference simplex used in training, simulation evaluation, and hardware anchor tests.

\subsection{Hardware Evaluation Protocol}
\label{app:hardware_protocol}

Section~\ref{sec:real_robot} gives the shared hardware setup, preference anchors, and trial counts. Table~\ref{tab:hardware_protocols} distinguishes the three command regimes: replay provides the command-matched comparison, fast tests dynamic joystick-driven locomotion, and slow emphasizes low-speed operation with near-standstill transitions.

\begin{figure*}[!t]
\centering
\includegraphics[
    height=0.97\textheight,
    keepaspectratio
]{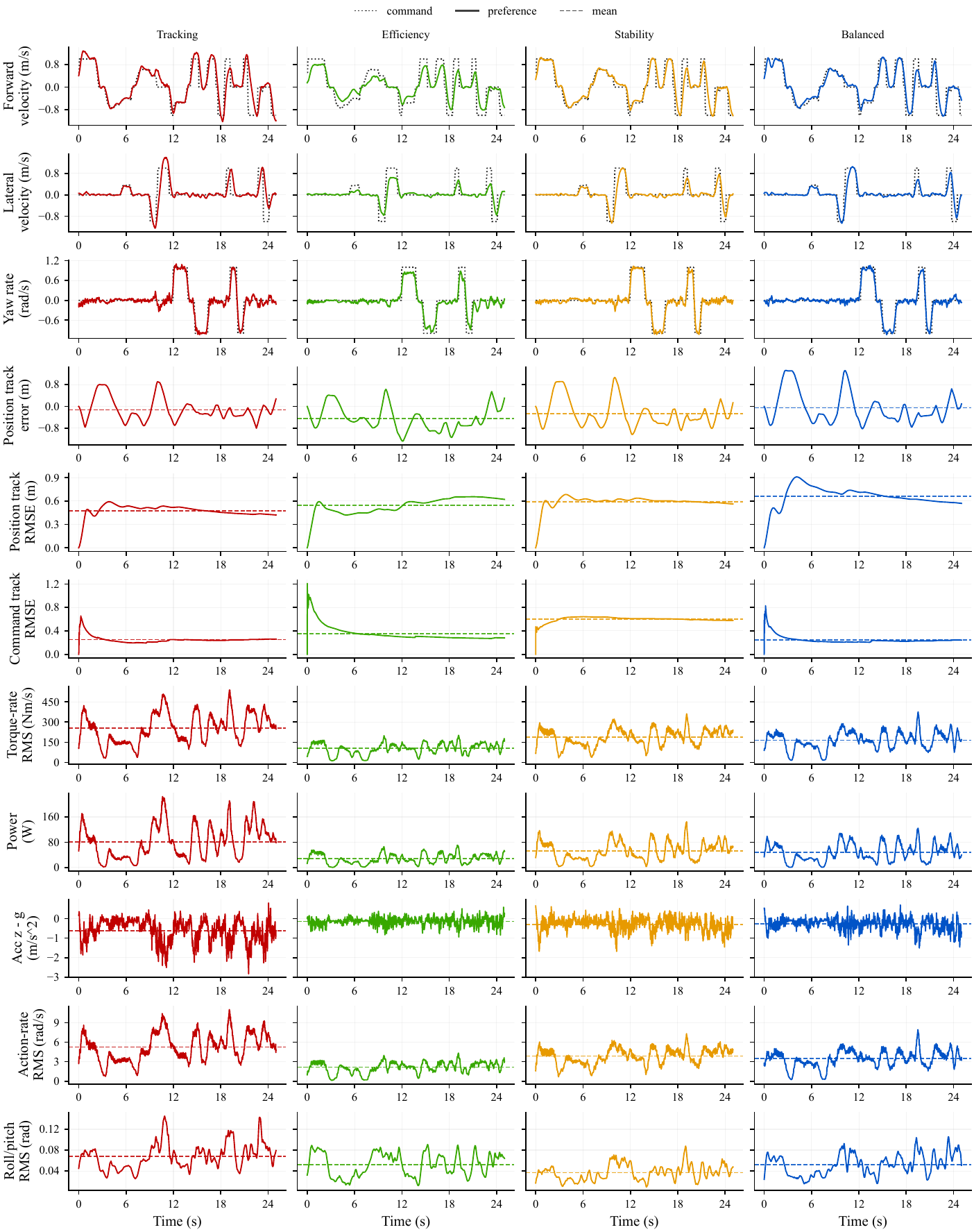}
\caption{\textbf{Replay command grid.} Identical recorded command sequence is replayed for all preference anchors, isolating preference-induced differences in trajectory tracking, energetic cost, and body regulation under matched command conditions.}
\label{fig:supp_hw_grid_replay}
\end{figure*}

Figure~\ref{fig:supp_hw_grid_replay} provides the detailed command-matched diagnostic for the Replay regime. Because all preferences receive the same recorded command sequence, differences in trajectory tracking, energy use, and body regulation are attributable to the deployed semantic preference rather than operator-input variation.

\begin{table}[t]
\caption{Real-robot evaluation regimes.}
\label{tab:hardware_protocols}
\centering
\footnotesize
\setlength{\tabcolsep}{5pt}
\begin{tabular}{lll}
\toprule
Regime & Command protocol & Evaluation focus \\
\midrule
Replay & Recorded, identical & Command-matched comparison \\
Fast   & Manual joystick    & Dynamic locomotion \\
Slow   & Manual joystick    & Near-standstill locomotion \\
\bottomrule
\end{tabular}
\end{table}


\section{Design Analysis and Ablations}
\label{app:analysis}
\label{sec:ablations}

This appendix analyzes the design choices behind \ac{promo}: separating semantic objectives from fixed locomotion priors, scalarizing before the clipped \ac{ppo} surrogate, and regularizing preference responsiveness at the action level.

\subsection{Objective/Prior Factorization}

\begin{proposition}[Objective/prior separation]
\label{prop:sep}
The aggregate locomotion-prior term does not define a preference axis. Specifically, the preference vector $w_t$ does not weight or scalarize $J_{\mathrm{prior}}$, although $J_{\mathrm{prior}}(\pi_\theta,w_t)$ may change indirectly through the preference-dependent state visitation distribution.
\end{proposition}

Only tracking, stability, and efficiency define the deployment interface and receive value heads, Pareto analysis, and controllability evaluation. The prior remains a fixed regularizer. All \ac{hv} values use these three outcomes and the common within-study normalization in Section~\ref{sec:experimental_evaluation}.

Table~\ref{tab:objprior_ablation} tests four treatments of the locomotion priors. We report mean reward and episode length instead of utility and \ac{cvar}10 because the fourth-objective variant changes both the utility definition and the preference dimension. 

Keeping the priors fixed and outside the semantic objectives gives the selected balance of controllability and reliability. Moving priors inside the semantic objectives slows learning and nearly triples the fall rate ($0.060\rightarrow0.165$), while removing priors increases hypervolume but reduces controllability and produces non-viable efficiency-heavy behavior. Exposing locomotion as a fourth semantic objective gives high aggregate utility but degrades controllability and fall rate. Across variants, aggregate metrics can improve while the operator-facing interface degrades.

\begin{table}[t]
\caption{Effect of objective/prior factorization on comparable training and interface metrics.}
\label{tab:objprior_ablation}
\centering
\footnotesize
\setlength{\tabcolsep}{4pt}
\resizebox{\columnwidth}{!}{%
\begin{tabular}{lccccc}
\toprule
Variant & Reward $\uparrow$ & Ep. len. $\uparrow$ & Ctrl. $\uparrow$ & Fall $\downarrow$ & \ac{hv} $\uparrow$ \\
\midrule
Fixed priors (ours) & $\mathbf{27.06}$ & $\mathbf{997.56}$ & $\mathbf{0.741}$ & $\mathbf{0.060}$ & $0.693$ \\
Prior removed & $25.63$ & $956.54$ & $0.638$ & $0.131$ & $\mathbf{0.906}$ \\
Priors inside objectives & $26.72$ & $996.60$ & $0.699$ & $0.165$ & $0.626$ \\
Prior as 4th objective & $26.86$ & $996.72$ & $0.447$ & $0.218$ & $0.512$ \\
\bottomrule
\end{tabular}
}
\end{table}

\begin{figure}[t]
    \centering
    \subfloat[Mean reward]{
        \includegraphics[width=0.75\linewidth]{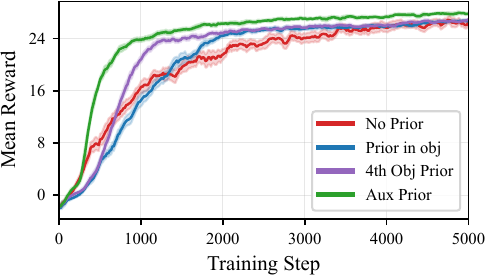}
    }\\[0.5em]
    \subfloat[Mean episode length]{
        \includegraphics[width=0.75\linewidth]{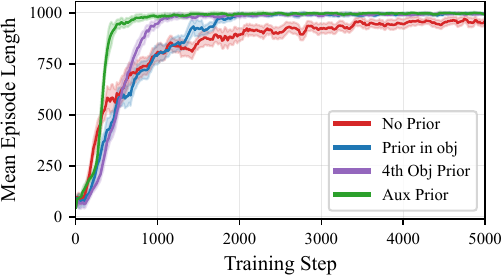}
    }
    \caption{Training behavior under alternative objective/prior formulations. Keeping the locomotion priors as fixed auxiliary shaping reaches stable locomotion and high reward fastest; embedding the priors in the semantic objectives or exposing them as a fourth objective delays convergence.}
    \label{fig:objdesign_curves}
\end{figure}

\subsection{Scalarization Before \acs{ppo} Clipping}
\label{app:ppo_scalarization}
The clipped \ac{ppo} surrogate is nonlinear in the advantage. For a scalar advantage, define
\begin{equation}
L_\epsilon(A;\varrho_t)
=
\min\!\left(
\varrho_t A,\,
\operatorname{clip}(\varrho_t,1-\epsilon,1+\epsilon)A
\right).
\end{equation}
Semantic scalarization and clipping do not generally commute:
\begin{equation}
L_\epsilon\!\left(\sum_i w_{t,i}A_i;\varrho_t\right)
\neq
\sum_i w_{t,i}L_\epsilon(A_i;\varrho_t),
\end{equation}
because the clipped branch selected by $L_\epsilon$ depends on the sign and magnitude of the scalarized advantage. \ac{promo} therefore estimates semantic objective values $V_i(s_t,w_t)$, forms the preference-scalarized semantic baseline $V_w(s_t,w_t)=\sum_i w_{t,i}V_i(s_t,w_t)$, includes the fixed-locomotion-prior reward directly in the scalar actor residual, normalizes scalar advantages over the batch, and then applies the clipped \ac{ppo} surrogate. The fixed locomotion priors are not assigned critic heads.

\subsection{Diversity Regularization}

\begin{table*}[t]
\caption{Diversity-regularization ablation. Training metrics are final-window means; evaluation metrics are averaged over three seeds.}
\label{tab:supp_diversity}
\centering
\footnotesize
\setlength{\tabcolsep}{6pt}
\renewcommand{\arraystretch}{1.08}
\begin{tabular}{lccccccc}
\toprule
Type & $\lambda$ & Reward $\uparrow$ & Ep. len. $\uparrow$ &
Ctrl. $\uparrow$ & \ac{cvar}10 $\uparrow$ & Fall $\downarrow$ & \ac{hv} $\uparrow$ \\
\midrule
None & -- & 26.8 & 997.3 & 0.80 & 0.012 & 0.10 & 0.56 \\
\midrule
\multirow{5}{*}{Action}
& $10^{-6}$          & 26.8 & 996.9 & \textbf{0.88} & 0.012 & \textbf{0.07} & 0.55 \\
& $3{\times}10^{-6}$ & 26.6 & 994.9 & 0.78 & 0.012 & \textbf{0.07} & 0.74 \\
& $\mathbf{10^{-5}}$ & \textbf{26.9} & \textbf{997.8} & 0.79 & 0.012 & 0.08 & 0.66 \\
& $3{\times}10^{-5}$ & 26.8 & 996.4 & 0.86 & 0.011 & 0.11 & 0.59 \\
& $10^{-4}$          & 26.5 & 996.7 & 0.48 & 0.009 & 0.13 & \textbf{0.85} \\
\midrule
\multirow{5}{*}{Latent}
& $10^{-6}$          & 26.3 & 997.1 & 0.86 & 0.011 & 0.07 & 0.62 \\
& $3{\times}10^{-6}$ & 26.5 & 997.1 & 0.78 & 0.011 & 0.18 & 0.35 \\
& $10^{-5}$          & 26.4 & 997.6 & 0.79 & \textbf{0.012} & 0.08 & 0.55 \\
& $3{\times}10^{-5}$ & 26.0 & 997.2 & 0.65 & 0.011 & 0.12 & 0.65 \\
& $10^{-4}$          & 26.2 & 996.4 & 0.81 & 0.011 & 0.11 & 0.44 \\
\bottomrule
\end{tabular}
\vspace{0.8\baselineskip}

\includegraphics[width=0.99\textwidth]{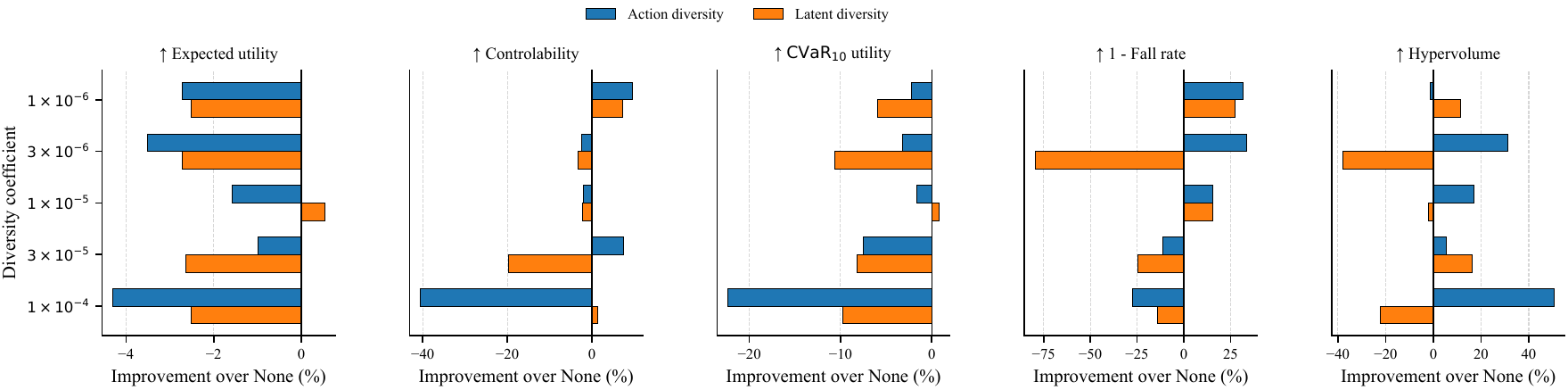}
\captionof{figure}{Sensitivity of deployment metrics to diversity regularization. Results are averaged over three seeds and shown as relative improvement over no diversity, exposing both metric-specific optima and coefficient sensitivity.}
\label{fig:div_eval_summary}
\end{table*}

Diversity regularization, in the spirit of diversity-regularized \ac{morl}~\cite{ambadkar2026d3po} and unsupervised skill discovery~\cite{eysenbach2019diayn}, is useful but coefficient-sensitive. We compare two mechanisms. Action diversity uses the training loss in~\eqref{eq:action_diversity_main}, directly separating the output distributions induced by different preferences at the same observation and detached history latent. The latent alternative regularizes an internal preference-conditioned representation $h_\theta(o_t,z_t^H,w_t)$,
\begin{equation}
\begin{split}
\mathcal{L}_{\mathrm{div}}^{\mathrm{lat}}
&=
\mathbb{E}
\Big[
\big(
\lVert h_\theta(o_t,z_t^H,w_t)-h_\theta(o_t,z_t^H,w'_t)\rVert \\
&\qquad
-
\alpha\lVert w_t-w'_t\rVert
\big)^2
\Big],
\end{split}
\label{eq:latent_diversity_app}
\end{equation}
where $\alpha$ sets the target scale between representation distance and preference distance. Action diversity acts on the deployed policy distribution itself, whereas latent diversity shapes the representation from which the policy is produced.

Table~\ref{tab:supp_diversity} summarizes the action/latent sweep, including reward and episode length over the final $10\%$ of training, while Fig.~\ref{fig:div_eval_summary} shows the corresponding improvement over no diversity. No coefficient dominates every metric: action diversity at $10^{-6}$ gives the largest controllability gain ($0.803\rightarrow0.879$) and a low fall rate ($0.067$), whereas the selected $10^{-5}$ setting broadens the Pareto spread ($0.657$ vs. $0.561$ hypervolume), reduces falls ($0.098\rightarrow0.083$), and slightly improves final reward and episode length. Latent diversity can improve individual metrics, but its response is less consistent across coefficients. Taken together, the sweep favors action-level regularization as the more reliable interface-level mechanism and shows that its coefficient should be selected using controllability, robustness, and Pareto coverage rather than training reward alone.


\section{Baseline Implementations}
\label{app:baseline_implementations}

All simulation baselines use the common flat-terrain evaluation used for \ac{promo}: a $12$-\ac{dof} quadruped with joint-position actions at $50$\,Hz, $20$\,s episodes, planar velocity commands in $[-1,1]$\,m/s, yaw-rate commands in $[-1,1]$\,rad/s, and matched termination conditions. All methods use on-policy \ac{ppo}-style training and are evaluated with the semantic decomposition in Appendix~\ref{app:rewards_preferences}.

\subsection{Single-Objective Locomotion Baselines}

\ac{tar} and \ac{rma} are fixed-trade-off controllers trained with the balanced scalar semantic reward plus fixed locomotion priors. Their deployment actors do not receive a preference input. \ac{tar} uses a history-based actor and aligns it with privileged critic information through auxiliary transition and velocity-prediction losses. \ac{rma} trains a privileged environment encoder and distills its latent into a history-based adaptation module for deployment.

\subsection{\acs{morl} Baselines}

\ac{moppo} uses a single preference-conditioned \ac{ppo} policy. The actor and critic receive a three-dimensional preference sampled from the same truncated simplex as \ac{promo} and held fixed throughout each rollout. The critic predicts one return per semantic objective, while the actor optimizes their preference-weighted combination together with the fixed locomotion priors.

\ac{dpmorl} uses a four-policy archive trained with fixed utilities: equal weighting and one utility emphasizing each of tracking, stability, and efficiency. Each policy observes the cumulative discounted semantic-objective return state and is trained with scalar \ac{ppo} on the incremental utility reward. The environment-interaction budget is divided equally across archive members, so each receives one quarter of the single full-budget run.

\subsection{\acs{promo}-\acs{sorl} Specialists}

\ac{promo}-\ac{sorl} comprises independently trained scalar policies using the \ac{promo} actor-critic architecture without preference conditioning, objective-wise semantic critics, or parameter sharing. The actor observes the same proprioceptive history used by \ac{promo} but receives no preference vector. We train balanced, tracking-, stability-, and efficiency-focused specialists by varying the semantic coefficients while retaining fixed locomotion priors. This baseline compares one preference-conditioned \ac{promo} policy against separately trained fixed scalarizations.

\end{document}